\documentclass[10pt,twocolumn,letterpaper]{article}

\usepackage[pagenumbers]{cvpr}

\usepackage{afterpage}
\usepackage[linesnumbered, ruled]{algorithm2e}
\usepackage{amsmath}
\usepackage{amssymb}
\usepackage{array}
\usepackage{booktabs}
\usepackage{caption}
\usepackage{enumitem}
\usepackage{graphicx}
\usepackage{microtype}
\usepackage{multirow}
\usepackage{stfloats}

\let\temp\paragraph
\renewcommand{\paragraph}[1]{\vspace{-2ex}\temp{#1.}}

\usepackage[pagebackref,breaklinks,colorlinks]{hyperref}
\usepackage[capitalize]{cleveref}

\def\cvprPaperID{337}
\def\confName{3DV}
\def\confYear{2024}

\SetKw{KwFor}{for}
\RestyleAlgo{ruled}

\begin{document}

\title{SlimmeRF: Slimmable Radiance Fields}

\author{Shiran Yuan\textsuperscript{\textrm{1,2,3,}}\thanks{Research done during internship with AIR.}\quad Hao Zhao\textsuperscript{\textrm{1,}}\thanks{Corresponding author.} \\
\textsuperscript{\textrm{1}}AIR, Tsinghua University\quad \textsuperscript{\textrm{2}}Duke University\quad \textsuperscript{\textrm{3}}Duke Kunshan University \\
{\tt\small sy250@duke.edu}, {\tt\small zhaohao@air.tsinghua.edu.cn}}
\maketitle

\begin{abstract}
Neural Radiance Field (NeRF) and its variants have recently emerged as successful methods for novel view synthesis and 3D scene reconstruction. However, most current NeRF models either achieve high accuracy using large model sizes, or achieve high memory-efficiency by trading off accuracy. This limits the applicable scope of any single model, since high-accuracy models might not fit in low-memory devices, and memory-efficient models might not satisfy high-quality requirements. To this end, we present SlimmeRF, a model that allows for instant \textbf{test-time} trade-offs between model size and accuracy through slimming, thus making the model simultaneously suitable for scenarios with different computing budgets. We achieve this through a newly proposed algorithm named Tensorial Rank Incrementation (TRaIn) which increases the rank of the model's tensorial representation gradually during training. We also observe that our model allows for more effective trade-offs in sparse-view scenarios, at times even achieving \textbf{higher} accuracy after being slimmed. We credit this to the fact that erroneous information such as floaters tend to be stored in components corresponding to higher ranks. Our implementation is available at \url{https://github.com/Shiran-Yuan/SlimmeRF}.
\end{abstract}%

\vspace{-4ex}

\section{Introduction}
\begin{figure}[t]
\centering
\vspace{-2ex}
\includegraphics[width=.46\textwidth]{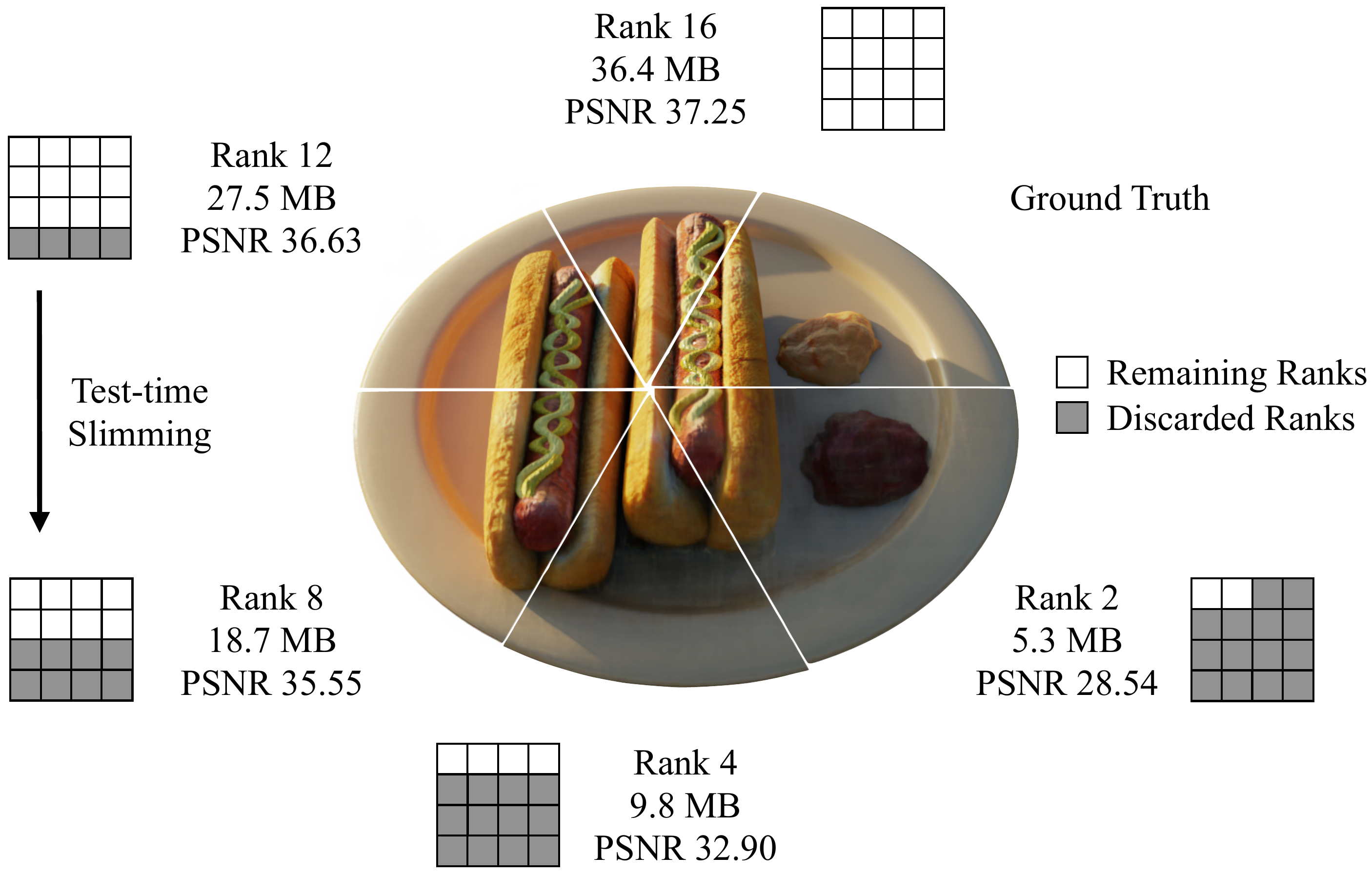}
\vspace{-2ex}
\caption[width=.5\textwidth]{\textbf{A single model, trained only once, could achieve multiple compression levels at test time.} Our SlimmeRF model enables trade-offs between model size and accuracy to be readily made at test time while not requiring any re-training. Shown in the figure are testing results for the \emph{Hotdog} scene at different compression levels, all from a single SlimmeRF-16 model. }
\label{fig1}
\vspace{-4ex}
\end{figure}%

View synthesis and the reconstruction of 3D scenes are longstanding problems in computer vision that have important applications in many domains. The NeRF~\cite{mildenhall2021} model, which represents the geometry and appearance of 3D scenes as neural fields, has recently proven to be an effective solution to this problem, allowing for high-quality 2D view synthesis and 3D geometric reconstruction. Based on this work, numerous variants were proposed, using novel model architectures to further improve NeRF's reconstruction quality, efficiency and applicability to different settings.

Despite the success of NeRF models, most still suffer from a common disadvantage that hinders flexibility: models trained in scenarios with loose requirements might be inapplicable in cases where requirements are stricter. Hence, if we target for models that are usable across different usage cases, we can only train them according to the strictest requirements, sacrificing accuracy. 

To tackle this issue, we propose SlimmeRF, a model that could be trained using high capacities and achieves fairly good results on systems with lower capacities via \emph{test-time slimming}. While test-time slimming by partially discarding parameters from the model could easily be achieved on explicit NeRF models such as Instant-NGP~\cite{muller2022} and TensoRF~\cite{chen2022}, such naïve approaches have very poor accuracy (as demonstrated by our experiments with the baseline in Section \ref{sec4}). We, in contrast, successfully achieve this via the mathematical framework of low-rank tensor approximation.

\paragraph{Our Holy Grail: Slimmability in NeRF}

Slimmability is the ability of trained models to retain high-quality results when part of the model is removed, and therefore allow for flexible trade-offs between memory size and accuracy during test time. This concept was explored previously in the context of recognition networks and generative adversarial networks~\cite{hou2021, li2021, yu2019}, but our approach is fundamentally different from those works: to achieve slimmability, we utilize low-rank approximation to characterize the 3D scene as component tensors. During training, we initially set the number of component tensors (i.e., the rank of the tensor decomposition) to a low level, and only increment the rank when the existing components have already been trained well, a procedure which we formalize as the TRaIn (Tensorial Rank Incrementation) algorithm. Through this procedure, we assure that the most important information (\eg, basic geometrical outline, dominant colors, \etc) is aggregated in the first few components, and hence that removing the last few components would not cause large drops in accuracy. 

Notably, CCNeRF~\cite{tang2022} is a prior work which successfully achieves dynamic compression at test time, hence being similar to our work in terms of ultimate goals. However, we note that it cannot achieve state-of-the-art performance in terms of accuracy. This could be attributed to the fact that the decomposed components they use are \emph{heterogeneous}: with vector-based components storing essential information in a compact manner and matrix-based components being discardable. In contrast, our work's representation of components is \emph{homogeneous}: all individual components follow the same structure, and hence the model's learning ability is not affected. 

In addition, since floater noises generally do not adhere to low-rank structures, we expect them to be mainly present in the components corresponding to higher ranks. Thus, we test our method on sparse-view scenarios, which usually suffer from floaters due to insufficient multi-view constraints. Our experiments verify that though SlimmeRF notably does not incorporate knowledge of any properties that are specific to sparse views, it still achieves fairly good results in sparse-view settings. We also observe the interesting result that sometimes the accuracy of SlimmeRF in sparse-view scenarios will \emph{increase} as the model is slimmed, which seems a counter-intuitive result (achieving higher performance with smaller memory and less model parameters retained), but agrees with our hypothesis that floaters reside in components corresponding to higher ranks.

\paragraph{Contributions}
The following are the main contributions of our work:
\begin{itemize}[noitemsep, nolistsep]
\item We introduce the novel objective and research direction of slimmability to NeRF research, which could be useful for creating models that need to be flexible in their applicable scope while retaining high accuracy.
\item We constructively demonstrate how slimmability could be achieved in NeRFs via creating a model which could both be used as a memory-efficient model and a normal model with state-of-the-art reconstruction quality, also formalizing our methods as the TRaIn algorithm.
\item We demonstrate that our model allows for effective trade-offs between model size and accuracy under sparse input scenarios, which signifies that under our framework, we can aggregate erroneous information such as floaters separately from key information.
\end{itemize}%

\section{Related Works}

\subsection{Scene Representation with Radiance Fields}

Recently, models based on Neural Radiance Fields (NeRF)~\cite{mildenhall2021} have become popular in areas such as generative modelling~\cite{chan2021,chan2022,gu2021,lin2023,liu2023,melas2023,meng2021,niemeyer2021,poole2022,qian2023,schwarz2020,wu2023}, pose estimation~\cite{chng2022,jeong2021,lin2021,rosinol2022,sucar2021,wangz2021,yen2021,zhu2022}, human body modelling~\cite{corona2022,jiang2022,lir2022,peng2021a,peng2021b,shao2022,zhao2022,zheng2022}, mobile 3D rendering~\cite{chen2023,garbin2021,lic2022,park2021,reiser2021,yan2023,yariv2023}, and so on. In addition, numerous advancements in efficiency were made~\cite{garbin2021,hedman2021,lindell2021,liu2020,muller2022,reiser2021,yu2021b}, thus expanding the scope of NeRF-based models' capabilities.

\begin{figure}[t]
\includegraphics[width=.46\textwidth]{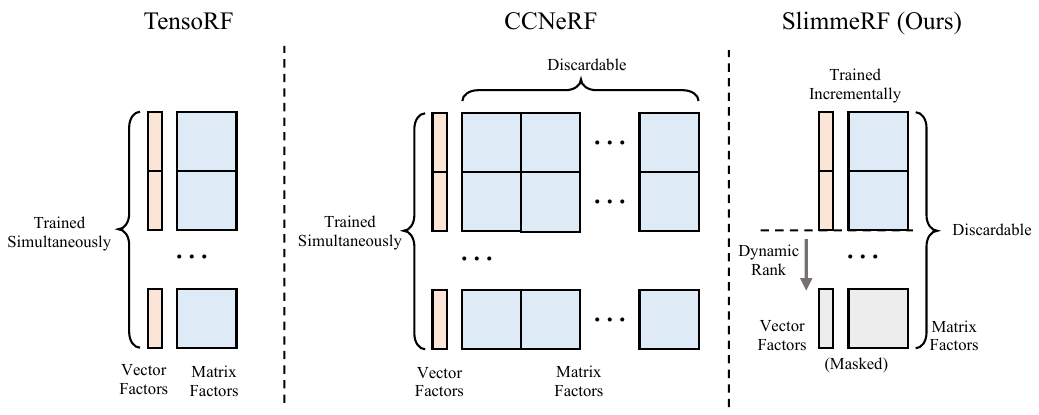}
\vspace{-2ex}
\caption[width=.5\textwidth]{\textbf{Comparison of our paradigm with similar models.} We compare the paradigm of SlimmeRF with TensoRF~\cite{chen2022} and CCNeRF~\cite{tang2022} to emphasize differences. In particular, note the difference between the heterogeneous paradigm of CCNeRF (only discarding matrix-based components) and the homogeneous paradigm of SlimmeRF (which discards components consisting of vector- and matrix- based components entirely). Also note that SlimmeRF is unique in its usage of incremental training.}
\label{fig2}
\vspace{-2ex}
\end{figure}%

A notable development in NeRF structures which lays the foundation for our work is the representation of scenes using explicit data structures~\cite{chen2022,garbin2021,hedman2021,liu2020,muller2022,reiser2021,yu2021a,yu2021b}. Not only does this allow for many improvements to the NeRF model, such as training and rendering speed, but it also makes the model parameters explainable. Our usage of tensorial grids represented by a Vector-Matrix Decomposition is also inspired by an important work in this direction, TensoRF~\cite{chen2022}.

\begin{figure*}[t]
\vspace{-2ex}
\centering
\includegraphics[width=\textwidth]{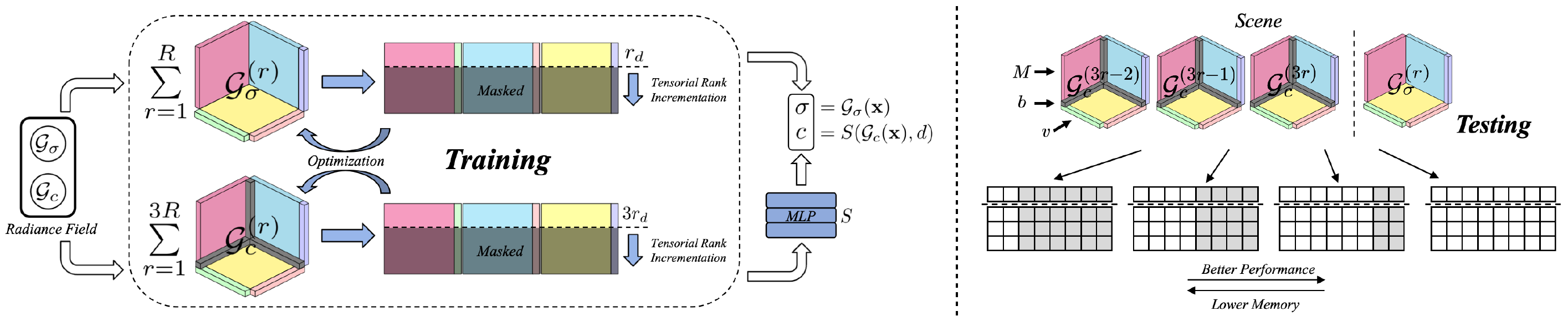}
\vspace{-4ex}
\caption[width=\textwidth]{\textbf{Illustration of SlimmeRF's model architecture.} Representing a radiance field as the tensorial grids $\mathcal{G}_\sigma$ and $\mathcal{G}_c$, we apply tensor decomposition to the grids to arrive at their low-rank approximation forms $\sum\limits_{r=1}^R\mathcal{G}_{\sigma}^{(r)}$ and $\sum\limits_{r=1}^{3R}\mathcal{G}_{c}^{(r)}$. They are then masked according to dynamic rank $r_d$ and trained through the Tensorial Rank Incrementation algorithm, with the grids being optimized using gradients calculated from the unmasked portions. Eventually the model predicts density $\sigma$ and appearance $c$ using the grids with respect to inputs position $\mathbf{x}$ and viewing direction $d$. The slimming procedure then discards elements corresponding to higher rank if lower memory usage is desired.}
\vspace{-4ex}
\label{fig3}
\end{figure*}%

\subsection{Low-Rank Approximation of Tensors}

The low-rank approximation of tensors is a mathematical technique that could be applied to compress data by exploiting the inherent low-rank properties of natural signals. It has seen a considerable number of applications in areas such as hyper-spectral image processing~\cite{pengs2021}, low-level vision~\cite{yokota2022}, and big data processing~\cite{song2019}. 

Though some previous works~\cite{chen2022,han2023,jang2022,shao2023,tang2022} have incorporated the concept of low-rank approximation or similar tensor decomposition-based concepts into NeRF, they neglect the potentials of tensors as slimmable representations. TensoRF~\cite{chen2022}, for instance, uses a tensorial structure for data representation but does not explore low-rank approximation and its applications; CCNeRF~\cite{tang2022}, as another example, explicitly states its relationship with low-rank tensor approximation, but it achieves so using a heterogeneous data structure where there are two inherently different types of factors, and test-time trade-offs are achieved by discarding matrix-based factors. This added complexity to the representation structure violates Occam's razor, and hence, as supported by our experimental results, causes reconstruction quality to be suboptimal. Our method is centered around the mathematical framework of low-rank tensor approximation, and achieves this via a novel incremental training approach (the TRaIn algorithm), achieving both test-time dynamic accuracy-size trade-offs and high model accuracy. For a clearer display of the differences, please see Figure~\ref{fig2}.

Hence, our work is the first to incorporate low-rank approximation of tensors into NeRF, for the purpose of using the approximation model to successfully achieve dynamic reduction of the model size.

\subsection{Compressing NeRFs}

Current works allow for compression of NeRFs mainly by directly making the data structure memory-efficient~\cite{chen2022,reiser2023,rho2023,tang2022,yan2023} or by compressing trained models for efficient storage~\cite{deng2023,li2023,zhao2023}. However, currently there are no works besides CCNeRF~\cite{tang2022} which explicitly allow for test-time dynamic trade-offs to be made between model size and model accuracy, and its reconstruction quality is suboptimal. Our model solves this problem by achieving slimmability and retaining high accuracy while not being slimmed.

\subsection{Training NeRFs with Sparse Inputs}

Since large amounts of views with adequate quality might be difficult to obtain in many real-life scenarios, training NeRFs with only very sparse inputs has been a topic of interest. There are many methods in this direction of research~\cite{barron2021, chen2021, chibane2021, jain2021, niemeyer2022, yu2021a}, which reconstruct scenes with considerable quality through few ($<10$) views or even only one view. Most such methods rely on specific priors regarding inherent geometric properties of real-life scenes.

We observe that while SlimmeRF is not specialized for this type of tasks, its performance is still fairly good (even exceeding specialized sparse-view methods in some viewing directions, as shown in Figure \ref{fig9}), despite requiring small memory, being quick-to-train, not needing pretraining, and achieving excellent slimmability. This supports our hypothesis that floaters and other erroneous information are more likely to be stored in higher ranks. 

\section{Methodology}

The structure of our model is displayed in Figure~\ref{fig3}. We use the geometry and appearance grid tensors to represent the scene, and use the Vector-Matrix (VM) Decomposition to represent it as tensorial components (details are elaborated on in Subsection~\ref{sub3.1}). The model is trained via Tensorial Rank Incrementation (see Subsection~\ref{sub3.2}), and the resulting grids represent our radiance field. During training, the addition of new components is simulated using masks for efficiency, and during testing, components can be directly discarded using truncation in order to slim the model (see Subsection~\ref{sub3.3}), allowing dynamic test-time trade-offs between memory-efficiency and accuracy. We also provide a theoretical explanation shedding light on the method's mechanism, which is elaborated in Appendix \textcolor{red}{D}.

\subsection{Formulation and Representation}
\label{sub3.1}

The radiance field has two outputs for a 3D location $x$, namely density $\sigma$ and appearance $c$, which can be represented as:
\begin{equation}
\label{rf_form}
\sigma,c=\mathcal{G}_{\sigma}(\mathbf{x}),S(\mathcal{G}_c(\mathbf{x}),d)
\end{equation} %
where $S$ is a preselected function (for which we use an MLP) and $d$ represents the viewing direction. This involves the tensors $\mathcal{G}_{\sigma}\in\mathbb{R}^{I\times J\times K}$ (the geometry tensor) and $\mathcal{G}_c\in\mathbb{R}^{I\times J\times K\times 3}$ (the appearance tensor), which we decompose, following the framework of \cite{chen2022}. Using the VM (Vector-Matrix) Decomposition, we define the following as components of $\mathcal{G}_{\sigma}$ and $\mathcal{G}_c$:
\begin{equation}
\label{sigma_component}
\mathcal{G}_{\sigma}^{(r)}=\mathbf{v}^{X}_{\sigma,r}\circ\mathbf{M}_{\sigma,r}^{YZ}+\mathbf{v}^{Y}_{\sigma,r}\circ\mathbf{M}_{\sigma,r}^{XZ}+\mathbf{v}^{Z}_{\sigma,r}\circ\mathbf{M}_{\sigma,r}^{XY}
\end{equation}%
\\[-7ex]
\begin{align}
\label{c_component}
\mathcal{G}_c^{(r)}=\mathbf{v}^{X}_{c,r}\circ\mathbf{M}_{c,r}^{YZ}\circ\mathbf{b}_{3r-2}&+\mathbf{v}^{Y}_{c,r}\circ\mathbf{M}_{c,r}^{XZ}\circ\mathbf{b}_{3r-1}\\ \nonumber
&+\mathbf{v}^{Z}_{c,r}\circ\mathbf{M}_{c,r}^{XY}\circ\mathbf{b}_{3r}
\end{align}%

The above decomposes the tensors into vectors $\mathbf{v}$, matrices $\mathbf{M}$, and a matrix $\mathbf{B}$. Each tensor component itself has a fixed set of related values in the factor matrices and vectors (corresponding to the rank $r$ at which the component appears). The components are therefore independent of each other, and the resulting grids can be calculated as follows:
\begin{equation}
\label{tensor_model}
\mathcal{G}_{\sigma},\mathcal{G}_c=\sum\limits_{r=1}^{R_{\sigma}}\mathcal{G}_{\sigma}^{(r)},\sum\limits_{r=1}^{R_c}\mathcal{G}_{c}^{(r)}
\end{equation}%
where $R_c=cR_{\sigma}$ for an appearance grid with $c$ channels. In the rest of this paper, unless otherwise stated, $R$ refers to $R_{\sigma}$, and we use $c=3$ (RGB channels).

Our key observation is that different tensorial components are learned simultaneously in most existing models such as TensoRF. This does not make use of the independence between different components in the tensorial decomposed representation, and thus in the resulting trained model, the importance of different components are statistically equal. This hinders slimmability since the removal or adjustment of any component would severely impact the model as a whole.

\paragraph{Remark}
A straightforward way to make use of component independence is to group the parameters into blocks and optimize one block of parameters at a time. This method, known as Block Coordinate Descent (BCD), had been successfully applied to many other areas \cite{meshi2017asynchronous, zhao2017physics}. However, this method would not work as intended in our scenario. This is because our approximation of the tensors $\mathcal{G}_{\sigma}$ and $\mathcal{G}_{c}$ relies on the hypothesis that both of the tensors have low-rank structures. If only one block of components is trained at once, then the optimization objective for this block becomes learning the residual from other learned components. Residual learning has been applied in other scenarios such as CNN compression~\cite{guo2017}, but would impair model performance in our case as the residual is not guaranteed to be low-rank.

Inspired by BCD, we innovatively propose another method adapted to our specific task. The intuition behind our approach is that instead of imposing hard constraints on the components trained in each iteration, we gradually increase the learning power of the model and add new components to the model when needed. We call this approach \textbf{T}ensorial \textbf{Ra}nk \textbf{In}crementation, or TRaIn.

\subsection{Tensorial Rank Incrementation}
\label{sub3.2}

Our approach first limits the rank of the model, then increments the rank progressively when needed by adding more components. This is formally similar to the previously mentioned naïve BCD approach, but fundamentally different in that we incrementally add components to the model and train them collectively instead of optimizing separate blocks of parameters individually. Hence the connection between different components from the perspective of the model as a whole is not lost in the training process.

More specifically, parameters are divided into groups based on their relation to specific components, according to equations (\ref{sigma_component}) and (\ref{c_component}). Each time the model rank is incremented, another group of parameters is appended to the set of parameters being trained.

\paragraph{Principles of Rank Incrementation}
Implementation-wise, we use a dynamic rank $r_d$ to limit the model's learning capacity, replacing the maximum rank $R$ in (\ref{tensor_model}). As for the incrementation of $r_d$, an intuitive method would be to increment $r_d$ when the low-rank model converges, \ie when the loss changes at a low rate. However, tests show that it is difficult to determine convergence of the model since the loss is frequently unstable. Furthermore, waiting for the low-rank models to converge every time the dynamic rank changes would cause the converge of the entire model to be slowed. Hence, we counter-intuitively increment $r_d$ only when the loss changes at a \emph{high} rate instead, representing the phase where the model is learning fast and thus is also more able to make use of the extra learning power from incrementing the model dynamic rank. 

\paragraph{Criteria for Rank Incrementation}
Specifically, we use a hyper-parameter $\upsilon$ to control this process. For the rank incrementation process, we use the following MSE loss $L_{\textrm{MSE}}$, calculated from marching along and sampling points on rays:
\begin{equation}
\label{mse_loss}
L_{\textrm{MSE}}=\frac1R\sum\limits_{r=1}^R||c_r^*-\sum\limits_{n=1}^N(t_r^{(n)}(1-\textrm{e}^{-\delta_r\sigma_r^{(n)}})c_r^{(n)})||^2_2
\end{equation}%
where $N$ is the number of points sampled along each ray and $R$ is the number of sampled rays per batch; $c^*_r$ is the ground truth value of the pixel corresponding to the ray $r$; $\delta_r$ is the step size of ray $r$; $\sigma_r^{(n)}$ and $c_r^{(n)}$ respectively represent the predicted density and appearance of the ray $r$ at point $n$; $t^{(n)}_r$ represents the transmittance of the ray $r$ at point $n$, calculated by $t^{(n)}_r=\exp(-\sum\limits_{m=1}^{n-1}\delta_r\sigma_r^{(m)})$.

Let $L^{(i)}_{\textrm{MSE}}$ be the mean of MSE losses calculated from (\ref{mse_loss}) across all views on iteration $i$. We increment $r_d$ when the following inequality holds after training iteration $i$:
\begin{equation}
\label{rank_inc}
\frac{|L^{(i-1)}_{\textrm{MSE}}-L^{(i)}_{\textrm{MSE}}|}{L^{(i)}_{\textrm{MSE}}}>\upsilon
\end{equation}%

In experimentation with sparse inputs it was observed that sometimes the loss changes dramatically during the first iterations and causes the first few components to be learnt simultaneously. To prevent this, an additional optional constraint is added: that rank incrementations can only take place after a fixed amount $\eta$ of iterations has passed from the previous rank incrementation. However, in non-sparse input scenarios the performance often declines after introducing the $\eta$-constraint, and hence in those scenarios we set $\eta=0$. Ablation studies and parameter sensitivity analyses of $\upsilon$ and $\eta$ could be found in Table~\ref{tab3}.

\begin{figure*}[t]
\centering
\vspace{-2ex}
\includegraphics[width=\textwidth]{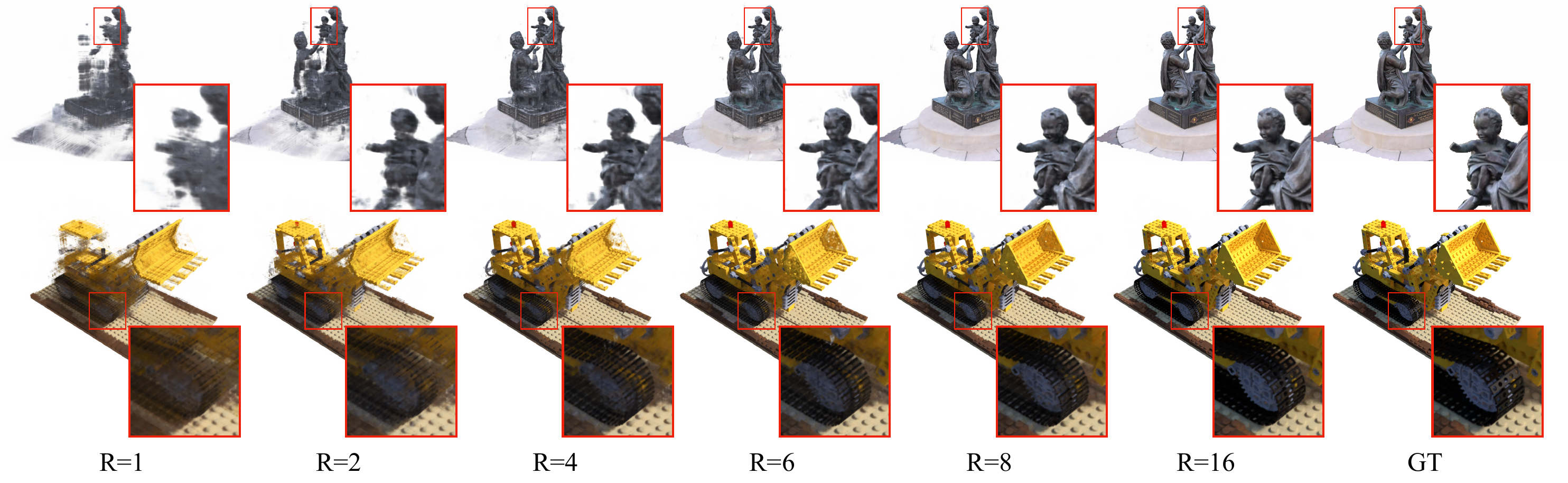}
\vspace{-4ex}
\caption[width=\textwidth]{\textbf{Automatic ``division of labor" between components.} As demonstrated by the testing results above (on \emph{Family} from Tanks \& Temples and \emph{Lego} from NeRF Synthetic), the lower-rank components of SlimmeRF prioritize learning general information regarding the scene such as structure and main color, and more delicate tasks such as texture patterns are handled by higher-rank components. Hence most of the important information are retained during model slimming, where components containing details are discarded, leaving fundamental knowledge regarding the scene intact.}
\vspace{-4ex}
\label{fig4}
\end{figure*}%

\paragraph{Intuition and Theory} 

The intuition behind our method is that when the model has low learning capacity, the components which are trained first will automatically prioritize learning essential information such as structure and main colors. Theoretical insights relating to this intuition are provided in Appendix \textcolor{red}{D}.

As verified empirically in Figure \ref{fig4}, this allows an automatic ``division of labor" to form in between the components. Components corresponding to lower ranks hold on to the vital information they acquired during training with low $r_d$, and components corresponding to higher ranks thus store more detailed information.

\subsection{Train-Time Masking and Test-Time Truncation}
\label{sub3.3}
\vspace{2ex}
\paragraph{Training Stage}

Changing the size of the parameter tensors during training time is a costly operation since all related data need to be renewed. Hence, we use masks to simulate truncation during training. Specifically, we replace the original parameters as follows:
\begin{align}
\label{v_masked}
\mathop{\oplus}\limits_{r=1}^{R}\mathbf{v}_{\sigma,r}^\alpha,\mathop{\oplus}\limits_{r=1}^{3R}\mathbf{v}_{c,r}^\alpha&=\mathbf{V}_\sigma^{\alpha *}\otimes\mathbf{V}_\sigma^{\alpha\mu},\mathbf{V}_c^{\alpha *}\otimes\mathbf{V}_c^{\alpha\mu} \\
\label{m_masked}
\mathop{\oplus}\limits_{r=1}^{R}\mathbf{M}_{\sigma,r}^\alpha,\mathop{\oplus}\limits_{r=1}^{3R}\mathbf{M}_{c,r}^\alpha&=\mathcal{M}_\sigma^{\alpha *}\otimes\mathcal{M}_\sigma^{\alpha\mu},\mathcal{M}_c^{\alpha *}\otimes\mathcal{M}_c^{\alpha\mu}
\end{align}%
where $\oplus$ is the stacking/concatenation operator and $\otimes$ is the Hadamard product; in (\ref{v_masked}), $\alpha\in\{X,Y,Z\}$ and in (\ref{m_masked}), $\alpha\in\{XY,YZ,XZ\}$. The matrix/tensor with the superscript $\mu$ is the mask, which we directly control in the algorithm, and the matrix/tensor with the superscript $*$ is learnable.

\begin{algorithm}[hb]
\small
Initialize model and parameters $\mathbf{v}$, $\mathbf{M}$, $\mathbf{B}$\;
Initialize masks according to (\ref{v_mask_sigma}), (\ref{v_mask_c}), (\ref{m_mask_sigma}), (\ref{m_mask_c})\;
Initialize $r_d$\;
$last\_inc\gets 1$\;
\For{$it\gets 1$ \KwTo $maxiter$}{
    Calculate $\mathbf{v}$ and $\mathbf{M}$ based on (\ref{v_masked}) and (\ref{m_masked})\;
    Calculate gradients with respect to $\mathbf{v}$ and $\mathbf{M}$\;
    Optimize $\mathbf{v}^*$ and $\mathbf{M}^*$ based on gradients\;
    Optimize the rest of the model normally\;
    \If{$it-last\_inc>\eta$}{
        Calculate $L_{\textrm{MSE}}$ from (\ref{mse_loss})\;
        \If{(\ref{rank_inc}) is satisfied}{
            Increment $r_d$\;
            $last\_inc\gets r_d$\;
            Update $\mathbf{V}^\mu$ and $\mathcal{M}^\mu$\;
        }
    }
}
\caption{The TRaIn Algorithm}
\label{alg}
\end{algorithm}%

\begin{table*}[tp]
\vspace{-6ex}
\footnotesize
\begin{center}
\begin{tabular}{|c|c|cccccccc|c|}
\hline
NeRF Synthetic & Size (MB) & Chair & Drums & Ficus & Hotdog & Lego & Materials & Mic & Ship & Avg. \\
\hline
Plenoxels~\cite{yu2021b} & 783 & 33.97 & 25.35 & 31.83 & 36.43 & 34.09 & 29.14 & 33.27 & 29.61 & 31.71 \\
DVGO~\cite{sun2022} & 206 & 34.06 & 25.40 & 32.59 & 36.77 & 34.65 & 29.58 & 33.18 & 29.04 & 31.91 \\
TensoRF-192~\cite{chen2022} & 71.9 & \textbf{35.80} & \textbf{26.01} & \textbf{34.11} & \textbf{37.57} & \textbf{36.54} & \textbf{30.09} & \underline{34.97} & \textbf{30.72} & \textbf{33.23} \\
SlimmeRF-16 (Ours) & \textbf{48.3} & \underline{35.74} & \underline{25.80} & \underline{34.03} & \underline{37.25} & \underline{36.45} & \underline{30.04} & \textbf{35.06} & \underline{30.58} & \underline{33.12} \\
\hline
Tanks \& Temples & & Barn & Caterpillar & Family & Ignatius & Truck & & & & Avg. \\
\hline
Plenoxels~\cite{yu2021b} & & 26.07 & 24.64 & 32.33 & 27.51 & 26.59 & & & & 27.43 \\
DVGO~\cite{sun2022} & & 27.01 & \underline{26.00} & 33.75 & 28.16 & \textbf{27.15} & & & & 28.41 \\
TensoRF-192~\cite{chen2022} & & \textbf{27.22} & \textbf{26.19} & \underline{33.92} & \underline{28.34} & \underline{27.14} & & & & \textbf{28.56} \\
SlimmeRF-16 (Ours) & & \underline{27.18} & \underline{26.00} & \textbf{34.03} & \textbf{28.46} & 26.94 & & & & \underline{28.52} \\
\hline
\end{tabular}
\end{center}%
\vspace{-4ex}
\caption{\textbf{Comparisons with state-of-the-arts.} As shown, our method is very memory-efficient even without slimming, and successfully displays state-of-the-art level accuracy. All values in the table except the ``Size" column are all PSNR values in dB.}
\vspace{-4ex}
\label{tab1}
\end{table*}%

The mask controls the rank incrementation process. Supposing that the function $\mathbf{1}(\mathcal{T})$ produces a tensor of the same dimensions as $\mathcal{T}$ but contains exclusively the element 1, and that $\boldsymbol{\epsilon}(\mathcal{T})$ produces one that contains exclusively the element $\epsilon$, the masks are calculated as:
{\allowdisplaybreaks
\begin{align}
\label{v_mask_sigma}
\mathbf{V}_{\sigma}^{\alpha\mu}&=(\mathop{\oplus}\limits_{r=1}^{r_d}\mathbf{1}(\mathbf{v}_{\sigma,r}^\alpha))\oplus(\mathop{\oplus}\limits_{r=r_d+1}^{R}\boldsymbol{\epsilon}(\mathbf{v}_{\sigma,r}^\alpha)) \\
\label{v_mask_c}
\mathbf{V}_{c}^{\alpha\mu}&=(\mathop{\oplus}\limits_{r=1}^{3r_d}\mathbf{1}(\mathbf{v}_{c,r}^\alpha))\oplus(\mathop{\oplus}\limits_{r=3r_d+1}^{3R}\boldsymbol{\epsilon}(\mathbf{v}_{c,r}^\alpha)) \\
\label{m_mask_sigma}
\mathcal{M}_{\sigma}^{\alpha\mu}&=(\mathop{\oplus}\limits_{r=1}^{r_d}\mathbf{1}(\mathbf{M}_{\sigma,r}^\alpha))\oplus(\mathop{\oplus}\limits_{r=r_d+1}^{R}\boldsymbol{\epsilon}(\mathbf{M}_{\sigma,r}^\alpha)) \\
\label{m_mask_c}
\mathcal{M}_{c}^{\alpha\mu}&=(\mathop{\oplus}\limits_{r=1}^{3r_d}\mathbf{1}(\mathbf{M}_{c,r}^\alpha))\oplus(\mathop{\oplus}\limits_{r=3r_d+1}^{3R}\boldsymbol{\epsilon}(\mathbf{M}_{c,r}^\alpha))
\end{align}%
}%
where $\epsilon$ is a small positive value. Note that here we use the $\epsilon$ tensor instead of the zero tensor because otherwise, values of 0 in the masked tensor would cause gradient descent to fail on masked elements of the trainable tensors (similar to death of ReLU neurons in deep neural networks). Eventually, $\mathbf{V}^*$ and $\mathcal{M}^*$ are updated based on gradients of the loss with respect to $(\oplus\mathbf{v})$ and $(\oplus\mathbf{M})$.

\begin{figure}[t]
\centering
\includegraphics[width=.45\textwidth]{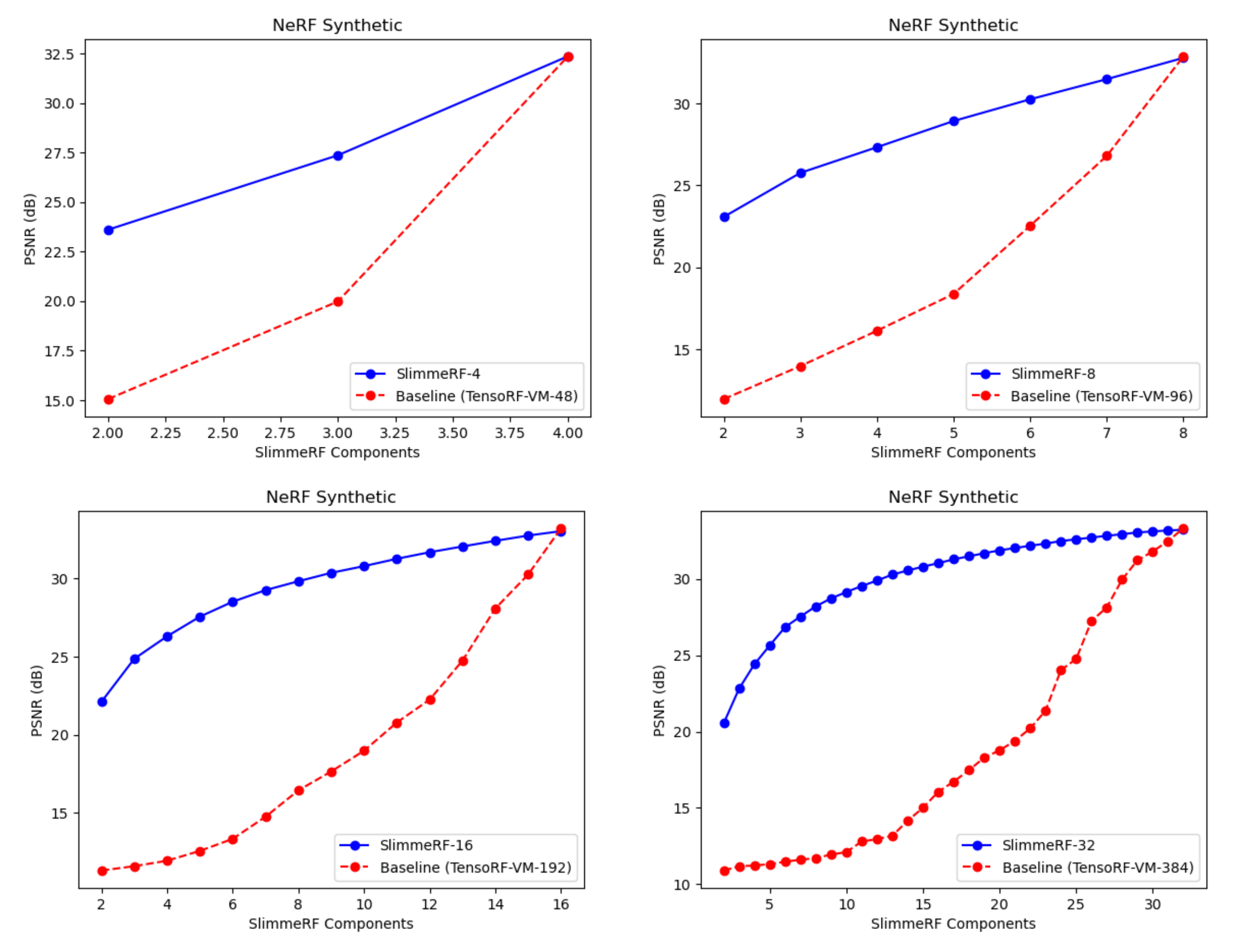}
\vspace{-2ex}
\caption[width=.5\textwidth]{\textbf{Comparisons with baselines.} A ``SlimmeRF component" corresponds to 12 TensoRF components since a set of tensorial components $\mathcal{G}_{\sigma}^{(r)}$, $\mathcal{G}_{c}^{(3r)}$, $\mathcal{G}_{c}^{(3r-1)}$, $\mathcal{G}_{c}^{(3r-2)}$ from SlimmeRF is equivalent to 12 matrix/vector components from TensoRF. Results obtained from the NeRF Synthetic dataset.}
\label{fig5}
\vspace{-4ex}
\end{figure}%

Hence, we arrive at the following representation for our radiance fields during training:
\begin{align}
\sigma,c&=[(\sum\limits_{r=1}^{r_d}\mathcal{G}_{\sigma}^{(r)})+\epsilon^2(\sum\limits_{r=r_d+1}^{R}\mathcal{G}_{\sigma}^{(r)})](\mathbf{x}),\\ \nonumber
&S([(\sum\limits_{r=1}^{3r_d}\mathcal{G}_{\sigma}^{(r)})+\epsilon^2(\sum\limits_{r=3r_d+1}^{3R}\mathcal{G}_{\sigma}^{(r)})](\mathbf{x}),d)
\end{align}%
with $r_d=R$ after the training is complete, hence again aligning with structures of (\ref{rf_form}) and (\ref{tensor_model}).

Other details in training are similar to TensoRF. We apply a coarse-to-fine training by shrinking the bounding box during training. The grids are also upsampled during fixed iterations via trilinear interpolation. Note that to save memory, float values are converted to 16-bit precision when the model is being saved and back to 32-bit precision afterwards. 

The formalized version of the TRaIn Algorithm is shown in Algorithm~\ref{alg}.

\paragraph{Testing Stage}

After the training process, however, masking becomes meaningless. Hence we truncate the component tensors during testing and save model in its slimmed form so that the model size could be reduced in cases where memory or efficiency constraints are important. Alternatively, the model could be saved in its non-slimmed form and slimmed before rendering according to applicational needs.

\begin{figure}[t]
\centering
\includegraphics[width=.4\textwidth]{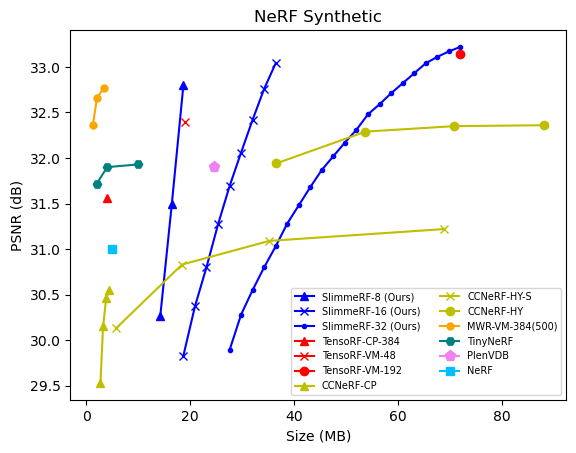}
\vspace{-2ex}
\caption[width=.5\textwidth]{\textbf{Comparison with compressive and memory-efficient methods.} We compare our model with other compressive and memory-efficient methods. As shown, our model could achieve high PSNR values during testing that are unobtainable via most other memory-efficient models. Also note that none of the models except CCNeRF support \textbf{test-time trade-offs} between size and accuracy, and thus all of those models are shown \emph{for reference} only rather than as competitors. Results obtained from the NeRF Synthetic dataset.}
\vspace{-4ex}
\label{fig6}
\end{figure}%

\section{Experiments}
\label{sec4}

In experiments, we use ``SlimmeRF-$R$" to refer to a model with rank $R$. More detailed per-scene testing results are available in the Appendices.

\paragraph{Experimental Details}
Our implementation is partially based on the codebase of TensoRF~\cite{chen2022}, uses the package PyTorch~\cite{paszke2019} for common modules, and applies the Adam optimizer~\cite{kingma2014} for optimization of the parameters. Experiments with NeRF Synthetic were conducted on a GeForce RTX 3090 GPU (24 GB), while experiments with LLFF and Tanks \& Temples were conducted on a Tesla V100 SXM2 GPU (32 GB). Regarding more detailed information such as hyper-parameter settings, please refer to Appendix A.

\begin{table*}[tp]
\vspace{-4ex}
\footnotesize
\begin{center}
\begin{tabular}{|c||c|c|c|c|c|c|c|c|}
\hline
 & \multicolumn{8}{c|}{Ours (SlimmeRF-24)} \\
\hline
 & R=2 & R=4 & R=6 & R=8 & R=12 & R=16 & R=20 & R=24 \\
\hline
3 Views & 15.51 & 16.63 & 16.73 & \textbf{16.76} & 16.75 & 16.74 & 16.73 & 16.72 \\
6 Views & 16.28 & 18.21 & 18.76 & 18.97 & 19.08 & \textbf{19.09} & \textbf{19.09} & 19.08 \\
9 Views & 17.09 & 19.66 & 20.51 & 20.97 & 21.21 & 21.29 & 21.34 & \textbf{21.35} \\
\hline
 & \multicolumn{8}{c|}{Specialized Sparse Input Models (For Reference)} \\
\hline
 & & SRF~\cite{chibane2021} & PixelNeRF~\cite{yu2021a} & MVSNeRF~\cite{chen2021} & mip-NeRF~\cite{barron2021} & DietNeRF~\cite{jain2021} & Reg-NeRF~\cite{niemeyer2022} & \\
\hline
3 Views & & 17.07 & 16.17 & 17.88 & 14.62 & 14.94 & 19.08 & \\
6 Views & & 16.75 & 17.03 & 19.99 & 20.87 & 21.75 & 23.10 & \\
9 Views & & 17.39 & 18.92 & 20.47 & 24.26 & 24.28 & 24.86 & \\
\hline
\end{tabular}%
\end{center}%
\vspace{-4ex}
\caption{\textbf{Quantitative results of sparse-input tests.} As shown, our model's slimmability is very high in sparse input cases, as the PSNR value barely decreases when the model is slimmed. Our method also performs comparably to specialized sparse-view models, despite not incorporating any specific priors for sparse-input scenarios. Note that while our model's performance does not excel state-of-the-arts, we mainly aim to test our model's slimmability rather than to compete with other models. All values in the table except the ``Size" column are all PSNR values in dB.}
\vspace{-4ex}
\label{tab2}
\end{table*}%

\paragraph{Comparison with TensoRF Baselines}

\begin{figure}[tp]
\vspace{-2ex}
\includegraphics[width=.5\textwidth]{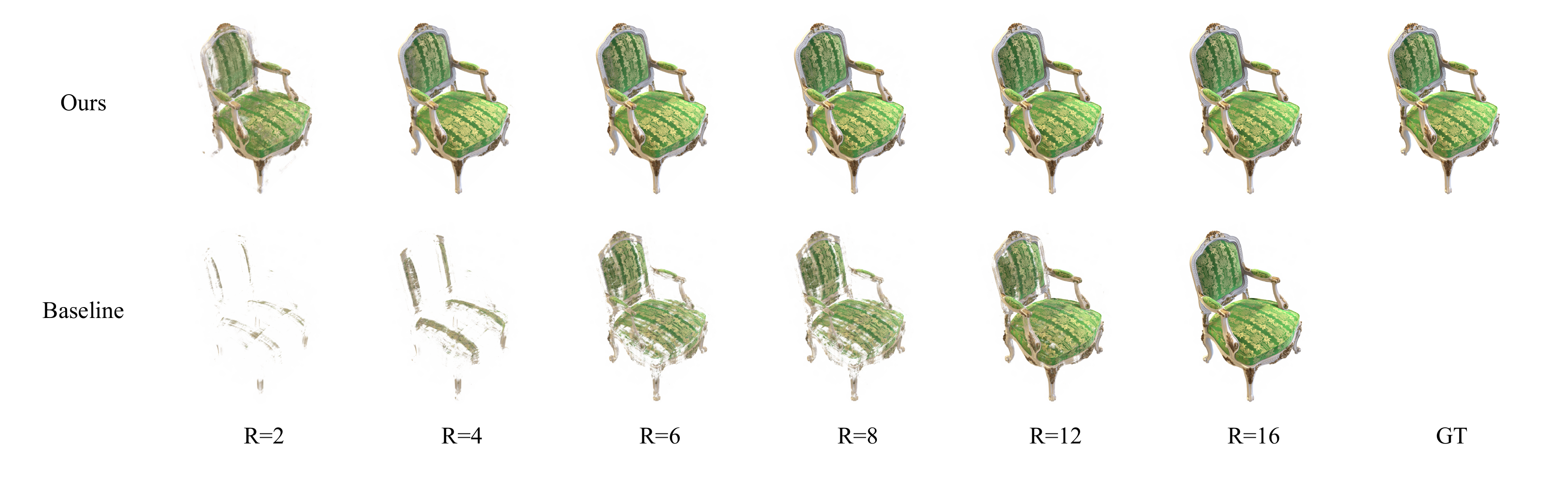}
\vspace{-4ex}
\caption[width=.5\textwidth]{\textbf{Qualitative comparison with baseline.} Shown is an experiment with the NeRF Synthetic scene \emph{Chair}. The model labelled ``ours" is a SlimmeRF-16, while the baseline is based on a TensoRF-VM-192. The value of ``$R$" represents the number of components left after slimming. }
\vspace{-4ex}
\label{fig7}
\end{figure}

Though our model structure is similar to TensoRF in terms of data representation, our theoretical premises are completely different from those of TensoRF's (low-rank tensor approximation versus tensor decomposition-based representation). Since we have already elaborated on the theoretical differences, we here demonstrate the differences between the two models in terms of empirical performance. We remove components from TensoRF in a similar fashion to our ``slimming" procedure, and proceed to normally calculate the PSNR values based on the slimmed TensoRF model as the baseline. Quantitative results of our experiment are shown in Figure~\ref{fig5}, and qualitative results are shown in Figure~\ref{fig7}. As shown, our model outperforms the baseline by a large margin, in accordance with theoretical analysis.

\begin{figure}[tp]
\vspace{-2ex}
\includegraphics[width=.48\textwidth]{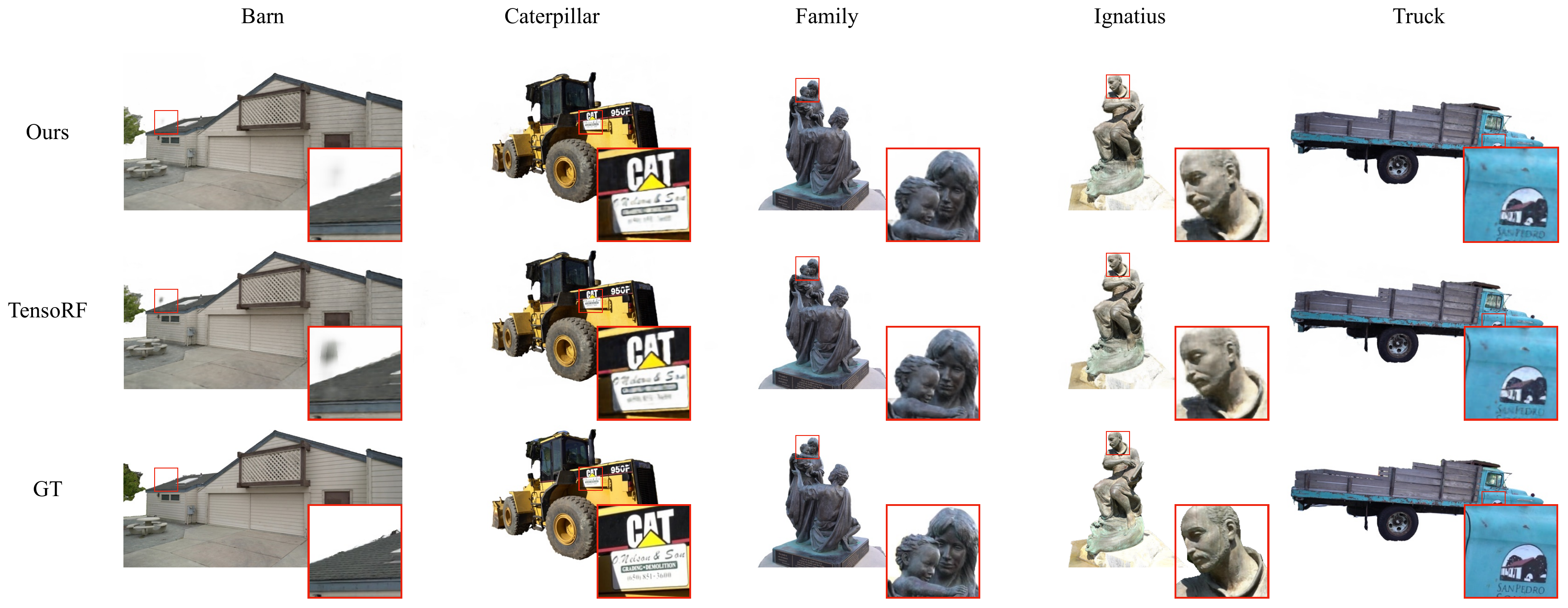}
\vspace{-4ex}
\caption[width=.5\textwidth]{\textbf{Qualitative comparisons with TensoRF.} The ``Ours" model is a SlimmeRF-16, and the TensoRF used is a TensoRF-VM-192. The scenes shown here are all from Tanks \& Temples~\cite{knapitsch2017}. As shown, our results are not inferior to TensoRF, and sometimes even exceed it. }
\label{fig8}
\vspace{-4ex}
\end{figure}

\paragraph{Comparison with State-of-the-Art Models}
We compare our model (without slimming) with the state-of-the-art models TensoRF~\cite{chen2022}, Plenoxels~\cite{fridovich2022}, and DVGO~\cite{sun2022} on the NeRF Synthetic dataset~\cite{mildenhall2021}. We note that while there exists models like mip-NeRF~\cite{barron2021} which can outperform the listed models, they take time on the magnitude of days to train, and so are not taken into our consideration. The results are shown in Table~\ref{tab1}. As shown, our model reaches performance comparable to state-of-the-art methods, and achieves slimmability while the baselines fails to. Figure~\ref{fig8} also shows a qualitative comparison of our results with TensoRF, which is representative of SOTA models.

\paragraph{Comparison with Compressive and Memory-Efficient Models}
We compare our model with the compressive and memory-efficient models TensoRF~\cite{chen2022}, CCNeRF~\cite{tang2022}, MWR (Masked Wavelet Representation)~\cite{rho2023}, TinyNeRF~\cite{zhao2023}, and PlenVDB~\cite{yan2023}. Note that we did not include works like Re:NeRF~\cite{deng2023} and VQRF~\cite{li2023} for comparison because they are post-training compression methods that can be used on models, and hence cannot be considered as models themselves. Note also that no works except CCNeRF support test-time trade-offs between model size and accuracy, and thus are only included for reference. Results are shown in Figure~\ref{fig6}. 

\begin{figure*}[tp]
\centering
\vspace{-6ex}
\includegraphics[width=.8\textwidth]{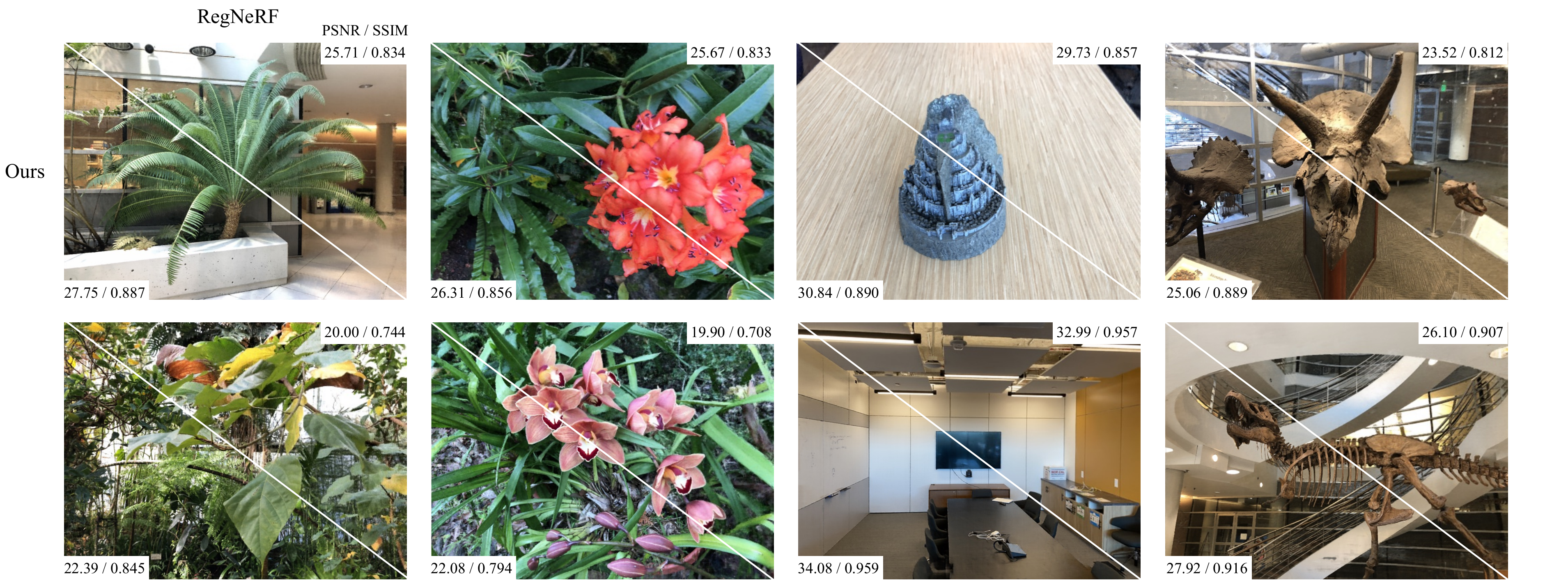}
\vspace{-2ex}
\caption[width=\textwidth]{\textbf{Qualitative comparison with Reg-NeRF.} Reg-NeRF~\cite{niemeyer2022} is a recent state-of-the-art model which performs well across many circumstances (also refer to Table~\ref{tab2} for quantitative information). Here we show some qualitative results of tests on LLFF (9 views) for both Reg-NeRF and our model (SlimmeRF-24). As displayed, in 2D synthesis tasks from certain viewing directions the performance of our model excels that of Reg-NeRF by a large margin. Because we do not attempt to hallucinate geometrical structures using prior assumptions, we did not achieve excelling results consistently across all viewing directions, but as this figure verifies, when enough knowledge associated with the specific viewing direction could be obtained, our performance could be superior to that of specialized sparse-view models. }
\label{fig9}
\end{figure*}%

\begin{table*}[tp]
\small
\begin{center}
\begin{tabular}{|c|cccccc|cccccc|}
\hline
 & \multicolumn{6}{c|}{$\upsilon$ (TRaIn) / Chair (NeRF Synthetic)} & \multicolumn{6}{c|}{$\eta$ ($\eta$-constraint) / Room (LLFF, 9 views)} \\
 & w/o & 0.1 & 0.2 & 0.3 & 0.4 & 0.5 & w/o & 50 & 100 & 200 & 400 & 600 \\
\hline
R=4  & 16.39 & 16.97 & 22.51 & \underline{24.03} & \textbf{30.11} & \multirow{4}{*}{N/A*} & 
11.70 & 23.60 & 23.21 & 23.59 & \underline{23.90} & \textbf{24.00} \\
R=8  & 19.89 & 20.11 & 30.25 & \underline{32.37} & \textbf{35.28} & & 
15.85 & 24.72 & 24.74 & 24.73 & \underline{24.98} & \textbf{25.01} \\
R=12 & 25.27 & 26.39 & 33.82 & \underline{34.74} & \textbf{35.29} & & 
22.57 & 25.18 & 25.36 & \textbf{25.42} & \underline{25.38} & 25.08 \\
R=16 & 35.80 & 35.81 & \textbf{35.82} & \textbf{35.82} & 35.34 & &
24.39 & 25.54 & \underline{25.64} & \textbf{25.86} & 25.41 & 25.06 \\
R=20 & & & & & & & 
25.08 & 25.75 & \underline{25.77} & \textbf{25.97} & 25.41 & 25.01 \\
R=24 & & & & & & & 
25.30 & 25.82 & \underline{25.86} & \textbf{25.98} & 25.40 & 25.06 \\
\hline
\end{tabular}
\end{center}
\vspace{-4ex}
\caption{\textbf{Ablation Studies and Parameter Sensitivity Analysis} The $\upsilon$ and $\eta$ parameters are subject to parameter sensitivity analysis. In addition, the cases where $\upsilon=0$ and $\eta=0$ can respectively be viewed as ablations of the TRaIn algorithm and of the $\eta$-constraint for sparse inputs. The experiments were respectively carried on \emph{Chair} from NeRF Synthetic with SlimmeRF-16 and \emph{Room} with 9 views from LLFF with SlimmeRF-24. \\
* The test with $\upsilon=0.5$ failed because (\ref{rank_inc}) was never satisfied, and the rank was not incremented.}
\vspace{-4ex}
\label{tab3}
\end{table*}%

\paragraph{Performance in Sparse-View Scenarios}
Our model's structure inherently stores data that do not conform to a low-rank structure into components of higher ranks. Hence, we hypothesized that this feature might be naturally beneficial to slimmability under sparse view reconstruction. 

To verify this, we trained a SlimmeRF-24 model on the LLFF dataset~\cite{mildenhall2019}. We then compare our model (and its slimmed versions) with the conditional models SRF~\cite{chibane2021}, PixelNeRF~\cite{yu2021a}, and MVSNeRF~\cite{chen2021}, and the unconditional models mip-NeRF~\cite{barron2021}, DietNeRF~\cite{jain2021}, and Reg-NeRF~\cite{niemeyer2022}, as references for the output quality of specialized sparse-view methods. 

We pretrain the conditional models on the DTU dataset~\cite{jensen2014} and then fine-tune them on LLFF for fair comparison (preventing overfitting during pretraining). Qualitative results from our model are shown in Figure~\ref{fig9}, and our model's quantitative performance is shown in Table~\ref{tab2}. 

We observe that our model's slimmability was greatly increased in sparse-view scenarios. Often a higher PSNR could be reached via discarding components. This is presumably due to the removal of floaters in components corresponding to higher ranks, and we include more empirical demonstrations of this as qualitative results in Appendix C. While our method did not achieve excelling results overall, we note that the performance of our model is superior to even specialized sparse-view models from some viewing directions, as shown by Figure~\ref{fig9}. 

\paragraph{Ablation Studies and Parameter Sensitivity Analysis}

We conduct ablation studies and parameter sensitivity analysis on the hyper-parameters $\upsilon$ and $\eta$. Results are displayed in Table~\ref{tab3}. As shown, when $\upsilon$ is small, the accuracy of the non-slimmed model will increase, at the expense of slimmability; slimmability increases with a large $\upsilon$ but might cause reduced model accuracy or even cause training to fail. 

As for $\eta$, in sparse-view scenarios, a small $\eta$ would produce both low-quality and non-slimmable results, similar to those of the TensoRF baseline. When $\eta$ is below the optimal value, both the quality and the slimmability of our model increase as $\eta$ increases, but when $\eta$ surpasses the optimal value the model's learning capability is severely restricted, and hence the quality of our model will decrease as $\eta$ increases while slimmability is not majorly affected. 

\paragraph{Comparison with BCD Baseline}

A baseline based on the naïve BCD method mentioned by Subsection \ref{sub3.1} was implemented and tested during our preliminary tests. During testing, the model's accuracy dramatically fluctuated in between the training of individual blocks. Hence, we failed to obtain any steady and reportable results from the baseline, reflecting the necessity to apply Tensorial Rank Incrementation instead of the intuitive approach of simply applying BCD to train the components separately in blocks.

\section{Conclusion}

In this work, we propose the new model SlimmeRF which allows for flexible and effective trade-offs between model accuracy and model size. We formalize this research objective as \emph{slimmability in NeRF}, and base our model on the newly introduced Tensorial Rank Incrementation algorithm to constructively show how this could be achieved. We also use experiments in sparse-input scenarios to demonstrate SlimmeRF's inherent property of storing essential and erroneous information (such as floaters) separately.

\section*{Acknowledgements}

This work is partially supported by AIR, Tsinghua University. The authors would like to thank the three anonymous reviewers for their recognition and suggestions. They would also like to thank (in alphabetical order) Huan'ang Gao, Liyi Luo, Yiyi Liao, and Zirui Wu for pertinent discussions and helpful feedback.

{\small
\bibliographystyle{ieee_fullname}

}

\end{document}


\title{SlimmeRF: Slimmable Radiance Fields\\\textnormal{Appendices}}

\author{Shiran Yuan\textsuperscript{\rm{1,2,3,}}\thanks{Research done during internship with AIR.}\quad Hao Zhao\textsuperscript{\rm{1,}}\thanks{Corresponding author.} \\
\textsuperscript{\rm{1}}AIR, Tsinghua University\quad \textsuperscript{\rm{2}}Duke University\quad \textsuperscript{\rm{3}}Duke Kunshan University \\
{\tt\small sy250@duke.edu}, {\tt\small zhaohao@air.tsinghua.edu.cn}}
\maketitle

The appendices shown here provide additional details for our implementation, testing results, and display more high-resolution qualitative comparisons. We also theoretically demonstrate the source of our model's slimmability. Please refer to \url{https://github.com/Shiran-Yuan/SlimmeRF} for our open-source repository.

\section{Additional Implementation Details}

All experiments with NeRF Synthetic~\cite{mildenhall2021} except those on \emph{Ficus} used the hyper-parameter settings $max\_iter=30000$, $\eta=0$, and $\upsilon=0.4$ (experiments on \emph{Ficus} used $\upsilon=0.2$ instead); all experiments with Tanks \& Temples~\cite{knapitsch2017} used the settings $max\_iter=30000$, $\eta=0$, and $\upsilon=0.2$; all experiments with LLFF~\cite{mildenhall2019} used the settings $max\_iter=30000$, $\eta=100$, and $\upsilon=0.1$. 

The MLP used for $S$ in \refeq{13} includes one hidden layer of size 128. The input and hidden layers use ReLU activation, and the output layer uses Sigmoid activation. We also use a coarse-to-fine training paradigm, with gradual grid upsampling and a bounding-box shrinking strategy. The grids are upsampled on iterations 2000, 3000, 4000, and 5500 for LLFF, and at iteration 7000 too for Synthetic NeRF and Tanks \& Temples. Bounding boxes are shrinked on iterations 2000 and 4000 for Synthetic NeRF and Tanks \& Temples, and on iteration 2500 for LLFF.

Some more specific details (such as adjustment terms to loss functions) are borrowed from the codebase of TensoRF~\cite{chen2022}, and we redirect the reader to their work or our open-source codebase for information regarding those.
 
\section{Additional Testing Results}

\subsection{Comparison with TensoRF Baselines}

We provide specific per-scene quantitative results for both SlimmeRF and the corresponding TensoRF baselines (with the equivalent number of tensorial components) across all 8 scenes from Synthetic NeRF and all 5 scenes from Tanks \& Temples. As shown, in all scenes our results very significantly exceed those of the baselines'.

The topmost row displays the number of remaining components after slimming. The leftmost column is the total number of components $R$ in SlimmeRF, corresponding to a baseline of TensoRF-VM-$12R$ (as explained in the caption of \figref{5}). 

Our results for Synthetic NeRF are displayed in Tables~\ref{chair}-\ref{synth}. Our results for Tanks \& Temples are displayed in Table~\ref{tnt}.

\subsection{Sparse Input Experiments}

We provide specific per-scene quantitative results for SlimmeRF-24's performance in LLFF with 3, 6, and 9 views respectively. Our results show that SlimmeRF achieves high slimmability in sparse-input cases. The results are displayed in Tables~\ref{LLFF1}-\ref{LLFF3}, and Table~\ref{LLFFavg} shows the averages.

\begin{table}[b]
\centering
\begin{tabular}{|c|c|c|c|}
\hline
\Aa & \Bb & \Cc & \multirow{2}{*}{GT} \\
\cline{1-3}
\Dd & \Ee & \Ff & \\
\hline
\end{tabular}
\caption{The arrangement of images in each group. The six images (as in the captions of Figures~\ref{llff1}-\ref{tnt2}) are labelled with letters \Aa-\Ff for indication of position.}
\label{arrange}
\end{table}

\section{Additional Qualitative Results}

We display high-resolution versions of our results achieved on LLFF (9 views), Synthetic NeRF, and Tanks \& Temples. Each group of selected results are displayed as 7 separate images, with the single image on the rightmost column being the Ground Truth. Each group of images is arranged as in Table~\ref{arrange}, and in the captions images are referred to according to the corresponding letters shown in the table.

\begin{figure*}[p]
\centering
\includegraphics[height=\figheight]{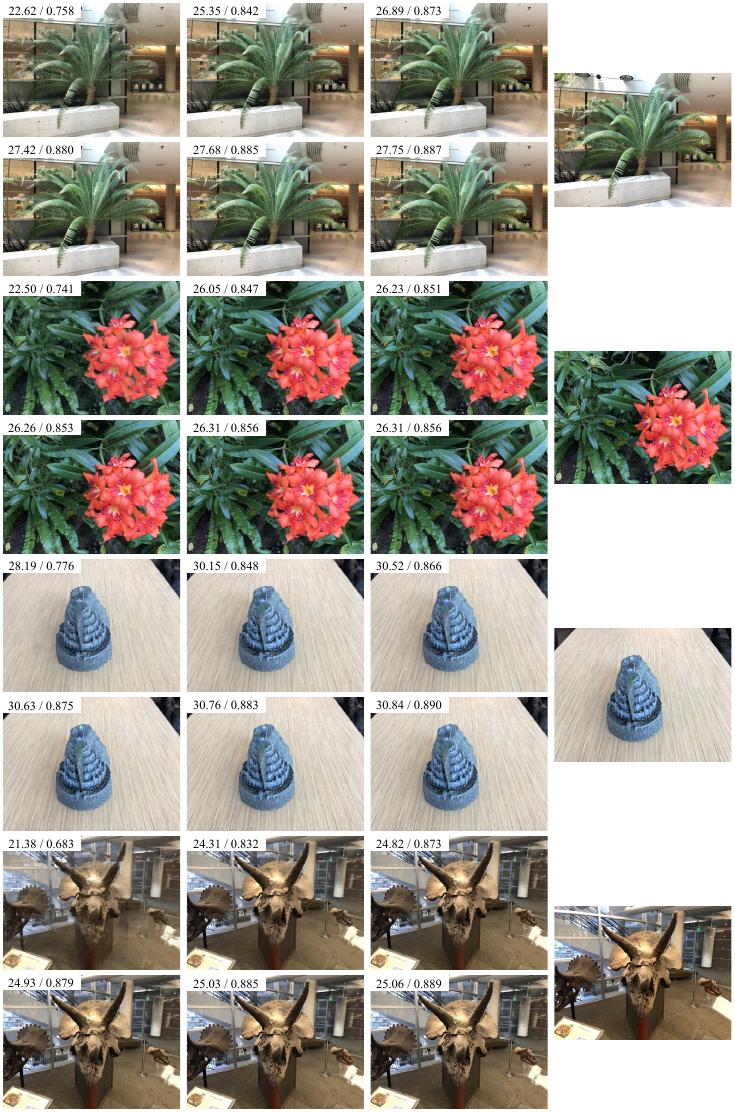}
\caption{Results for \emph{Fern}, \emph{Flower}, \emph{Fortress}, and \emph{Horns} of LLFF (9 Views). Correspondence between image position and the number of components left (see Table~\ref{arrange} for definition of the letters): \Aa 4; \Bb 8; \Cc 12; \Dd 16; \Ee 20; \Ff 24. The values displayed on the upper left corner are the PSNR value and the SSIM value.}
\label{llff1}
\end{figure*}

\begin{figure*}[p]
\centering
\includegraphics[height=\figheight]{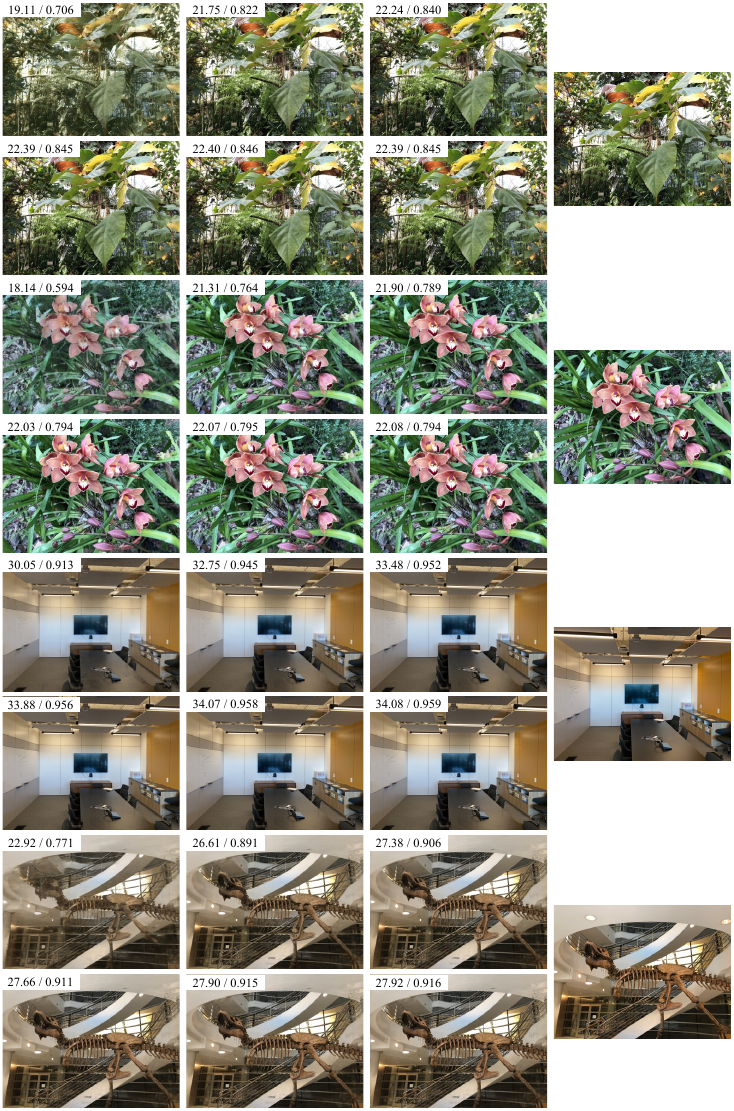}
\caption{Results for \emph{Leaves}, \emph{Orchids}, \emph{Room}, and \emph{T-Rex} of LLFF (9 Views). Correspondence between image position and the number of components left (see Table~\ref{arrange} for definition of the letters): \Aa 4; \Bb 8; \Cc 12; \Dd 16; \Ee 20; \Ff 24. The values displayed on the upper left corner are the PSNR value and the SSIM value.}
\label{llff2}
\end{figure*}

\begin{figure*}[p]
\centering
\includegraphics[height=\figheight]{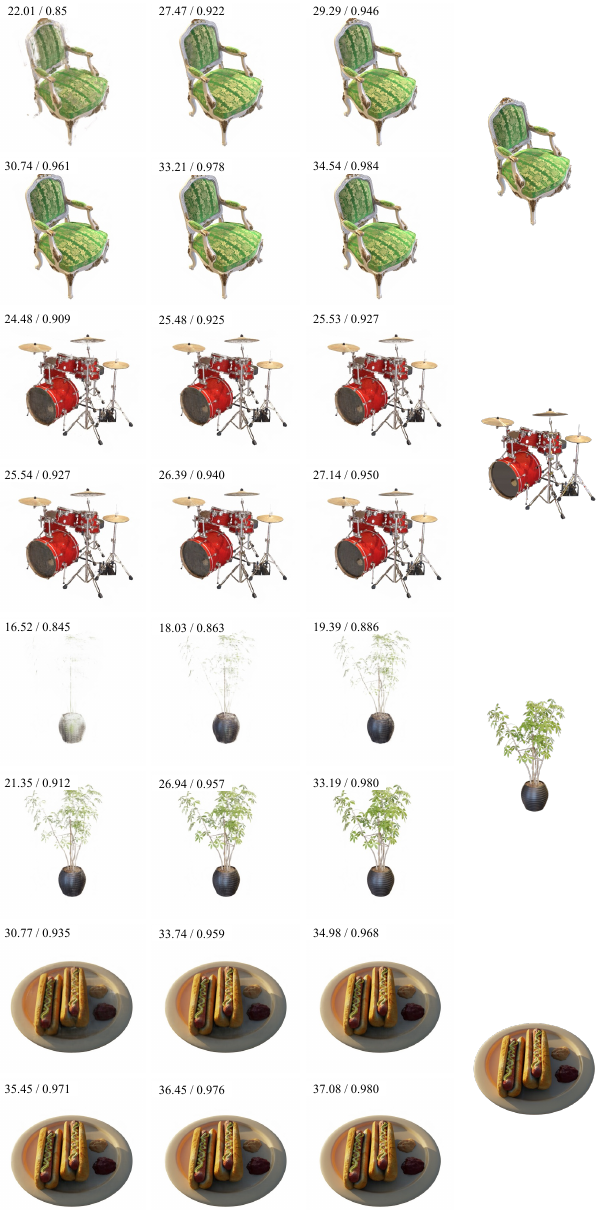}
\caption{Results for \emph{Chair}, \emph{Drums}, \emph{Ficus}, and \emph{Hotdog} of Synthetic NeRF. Correspondence between image position and the number of components left (see Table~\ref{arrange} for definition of the letters): \Aa 2; \Bb 4; \Cc 6; \Dd 8; \Ee 12; \Ff 16. The values displayed on the upper left corner are the PSNR value and the SSIM value.}
\label{synth1}
\end{figure*}

\begin{figure*}[p]
\centering
\includegraphics[height=\figheight]{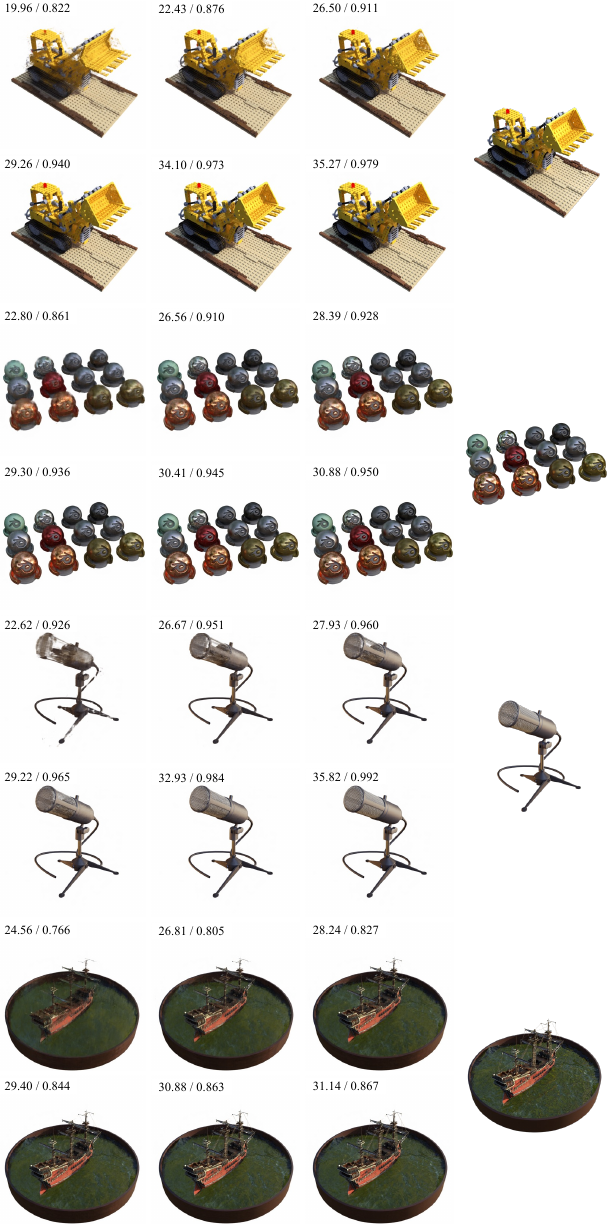}
\caption{Results for \emph{Lego}, \emph{Materials}, \emph{Mic}, and \emph{Ship} of Synthetic NeRF. Correspondence between image position and the number of components left (see Table~\ref{arrange} for definition of the letters): \Aa 2; \Bb 4; \Cc 6; \Dd 8; \Ee 12; \Ff 16. The values displayed on the upper left corner are the PSNR value and the SSIM value.}
\label{synth2}
\end{figure*}

\begin{figure*}[p]
\centering
\includegraphics[height=\figheight]{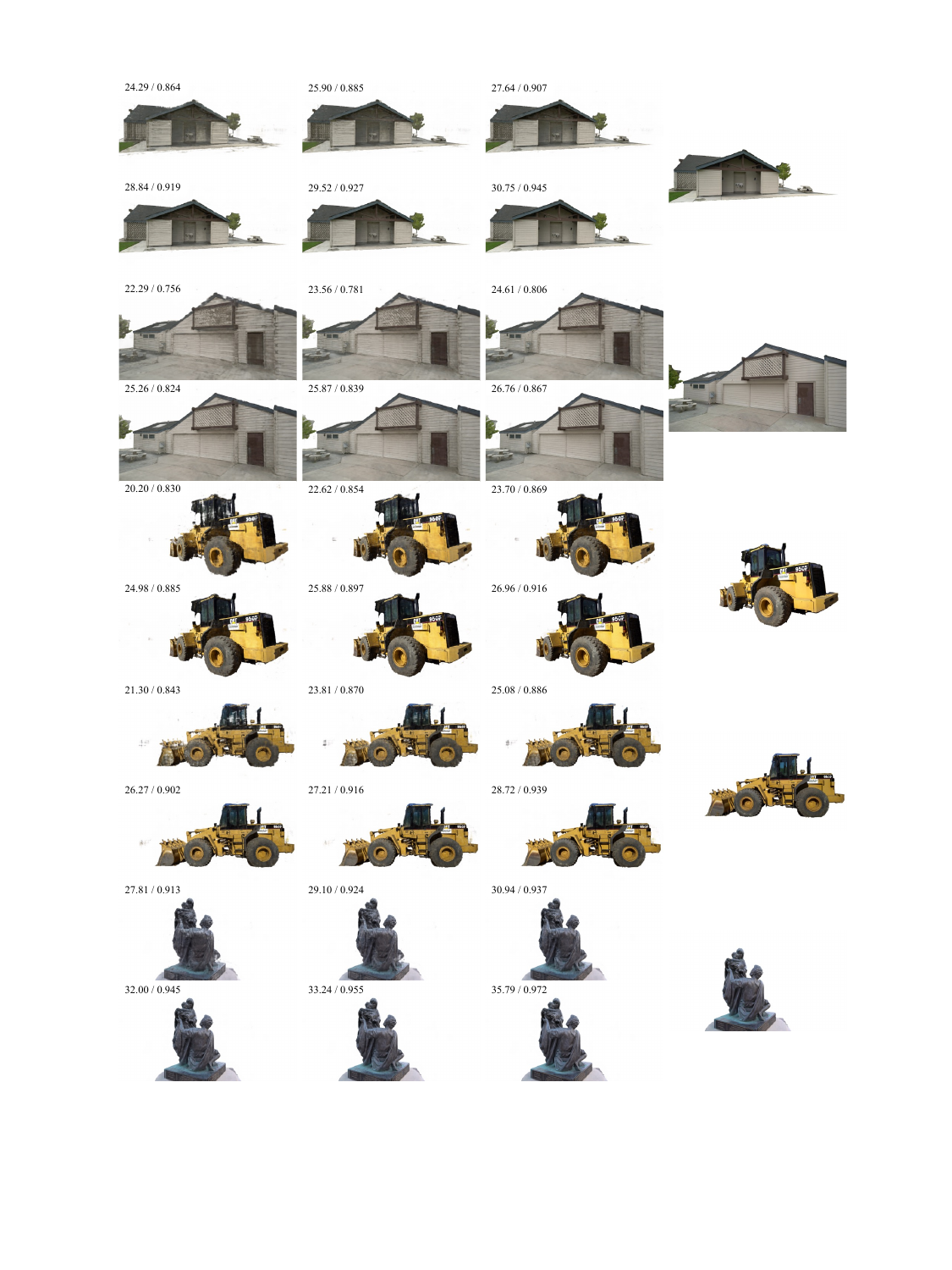}
\caption{Results for \emph{Barn}, \emph{Caterpillar}, and \emph{Family} of Tanks \& Temples. Correspondence between image position and the number of components left (see Table~\ref{arrange} for definition of the letters): \Aa 4; \Bb 6; \Cc 8; \Dd 10; \Ee 12; \Ff 16. The values displayed on the upper left corner are the PSNR value and the SSIM value.}
\label{tnt1}
\end{figure*}

\begin{figure*}[p]
\centering
\includegraphics[height=\figheight]{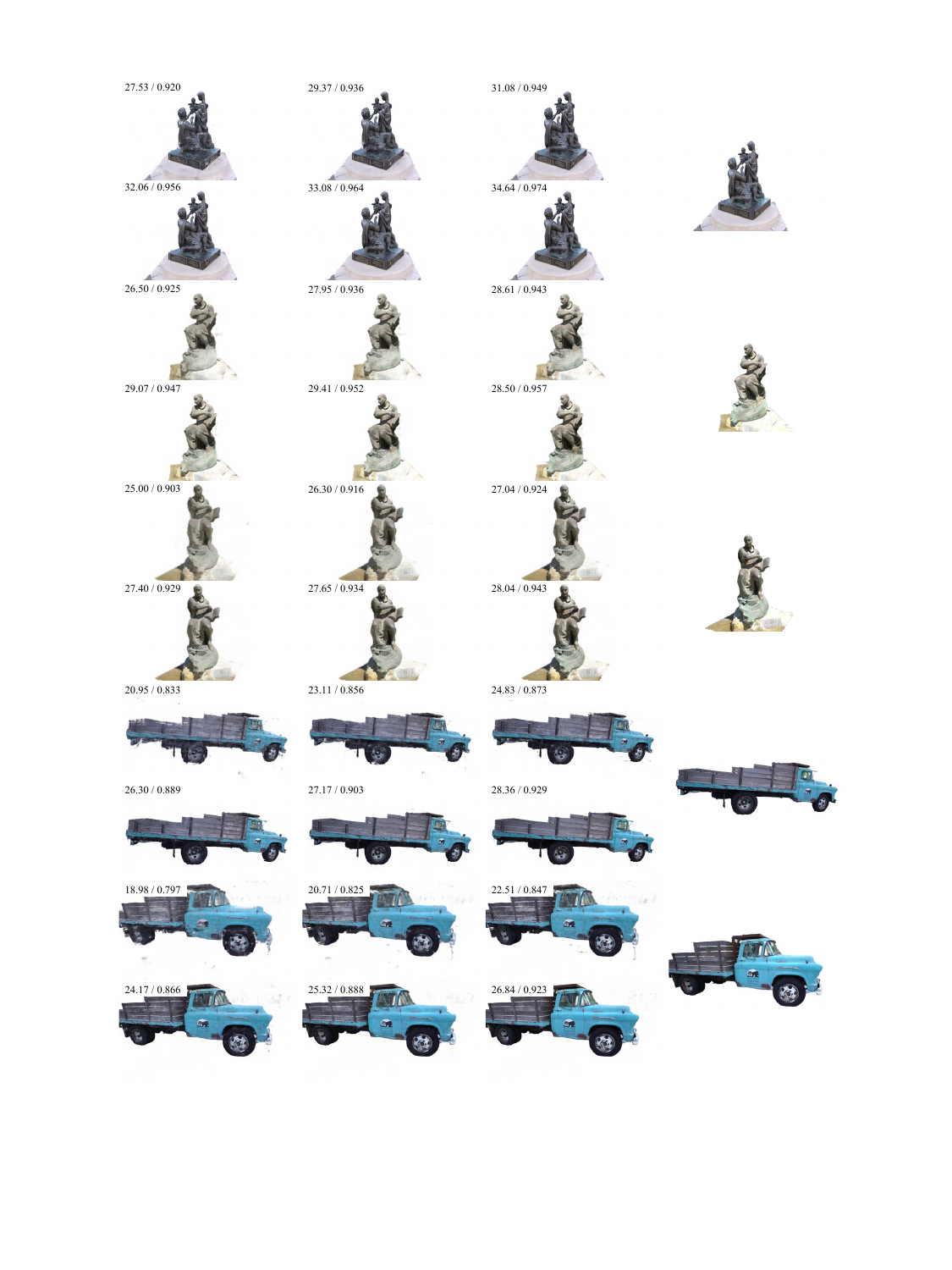}
\caption{Results for \emph{Family}, \emph{Ignatius}, and \emph{Truck} of Tanks \& Temples. Correspondence between image position and the number of components left (see Table~\ref{arrange} for definition of the letters): \Aa 4; \Bb 6; \Cc 8; \Dd 10; \Ee 12; \Ff 16. The values displayed on the upper left corner are the PSNR value and the SSIM value.}
\label{tnt2}
\end{figure*}

The results are shown in Figures~\ref{llff1}-\ref{tnt2}.

\section{Theoretical Mechanism}

In this section we mathematically demonstrate the mechanism behind the slimmability of our model. Specifically, we present a theoretical basis for the TRaIn algorithm's superiority in terms of slimmability over conventional simultaneous training paradigms. Our deduction reveals a theoretical upper bound on the partial derivative of the MSE loss with respect to elements of each tensorial component of the appearance grid, and demonstrates that utilization of the TRaIn algorithm instead of simultaneous training loosens this bound for tensorial components of lower rank. Hence this allows for tensorial components of lower rank to be trained faster under the TRaIn algorithm, thus achieving better slimmability.

\subsection{Preliminaries and Notation}
First, we rewrite~\refeq{5} as follows for clarity:
\begin{equation}
L_{\textrm{MSE}}(\mathcal{G}_{\sigma},\mathcal{G}_{c})=\frac1Q\sum\limits_{q=1}^Q(c_q^*-\sum\limits_{n=1}^Np_q^{(n)}c_q^{(n)})^2
\end{equation}%
where the original $r,R$ are respectively replaced by $q,Q$ to avoid confusion with ranks; $p_{q}^{(n)}$ is defined as follows:
\begin{equation}
p_{q}^{(n)}=\textrm{e}^{-\sum\limits_{m=1}^{n-1}\delta_q\sigma_q^{(m)}}(1-\textrm{e}^{-\delta_q\sigma_q^{(n)}})
\end{equation}%

We define the per-ray error functions $\theta^{(q)}$ as follows:
\begin{equation}
\theta^{(q)}(\mathcal{G}_{c})=(c_{q}^{*}-\sum\limits_{n=1}^Np_{q}^{(n)}c_{q}^{(n)})^2
\end{equation}%

We also use $\Omega$ to represent the set of all indices $\mathbf{i}$ in the tensorial grid $\mathcal{G}_{c}$.

\subsection{Linking Appearance and Components}
The MLP-represented function $S(m,d)$ modeling the appearance grid $\mathcal{G}_c$ satisfies the Lipschitz condition as a function of $m$, where $m=\mathcal{G}_{c}(\mathbf{x})$. Hence by definition there exists $K>0$ s.t. the following holds for any values of $m$ and $\epsilon$:
\begin{equation}
\label{s_bound}
|S(m,d)-S(m+\epsilon,d)|\leq K|\epsilon|
\end{equation}

A trivial consequence is as follows for all $m$:
\begin{equation}
\label{lipschitz}
-K\leq\frac{\partial S(m,d)}{\partial m}\leq K
\end{equation}

In addition, since $\mathcal{G}_c(\mathbf{x})$ is calculated via trilinear interpolation, we can create nonnegative tensors $\mathcal{Y}^{(n)}_q$ with the same dimensions as $\mathcal{G}_c$ according to the following:
\begin{align}
\label{bound_y}
\sum\limits_{\mathbf{i}\in\Omega}y^{(n)}_{q\mathbf{i}}=&1 \\
\mathcal{G}_{c}(\mathbf{x}^{(n)}_{q})=&\sum\limits_{\mathbf{i}\in\Omega}(y^{(n)}_{q\mathbf{i}}g_{c\mathbf{i}}^{(r)})
\end{align}

We can then compute the partial derivative of the appearance MLP output $c$ with respect to the component elements $g_{c\mathbf{i}}^{(r)}$ as follows:
\begin{align}
\label{c_to_component}
\frac{\partial c}{\partial g_{c\mathbf{i}}^{(r)}}=&\frac{\partial c}{\partial\sum\limits_{\mathbf{i}\in\Omega}y^{(n)}_{q\mathbf{i}}g_{c\mathbf{i}}}\frac{\partial\sum\limits_{\mathbf{i}\in\Omega}y^{(n)}_{q\mathbf{i}}g_{c\mathbf{i}}}{\partial g_{c\mathbf{i}}}\frac{\partial g_{c\mathbf{i}}}{\partial g_{c\mathbf{i}}^{(r)}} \\ \nonumber
=&y^{(n)}_{q\mathbf{i}}\frac{\partial S(\mathcal{G}_c(\mathbf{x}),d)}{\partial\sum\limits_{\mathbf{i}\in\Omega}y^{(n)}_{q\mathbf{i}}g_{c\mathbf{i}}} \\ \nonumber
=&y^{(n)}_{q\mathbf{i}}\frac{\partial S(\sum\limits_{\mathbf{i}\in\Omega}y^{(n)}_{q\mathbf{i}}g_{c\mathbf{i}},d)}{\partial\sum\limits_{\mathbf{i}\in\Omega}y^{(n)}_{q\mathbf{i}}g_{c\mathbf{i}}}
\end{align}

Hence we are able to deduce the following Lemma:
\begin{lemma}
\label{lemma_c}
The absolute value of the partial derivative of $c$ with respect to $g_{c\mathbf{i}}^{(r)}$ is theoretically upper bounded by the following relation:
\begin{equation}
|\frac{\partial c}{\partial g_{c\mathbf{i}}^{(r)}}|\leq K
\end{equation}
where $K$ is the Lipschitz constant of the appearance MLP $S(\mathcal{G}_c(\mathbf{x}),d)$ with respect to $\mathcal{G}_c(\mathbf{x})$.
\end{lemma}
\begin{proof}
Plugging (\ref{lipschitz}) into (\ref{c_to_component}), we have:
\begin{equation*}
-Ky^{(n)}_{q\mathbf{i}}\leq\frac{\partial c}{\partial g_{c\mathbf{i}}^{(r)}}\leq Ky^{(n)}_{q\mathbf{i}}
\end{equation*}
which can be combined with a trivial form of (\ref{bound_y}) to arrive at the Lemma.
\end{proof}

Note that it is also possible to find an upper bound to the Lipschitz constant $K$ such that this step of our proof could be made constructive~\cite{virmaux2018}. 

\subsection{Upper Bound on Learning Per-Ray Errors}
Define the function $f$ as follows:
\begin{equation}
\label{f_bound}
f(g_{c\mathbf{i}}^{(r)})=\sum\limits_{n=1}^N(p_{q}^{(n)}S(\sum\limits_{\mathbf{i}\in\Omega}y^{(n)}_{q\mathbf{i}}g_{c\mathbf{i}},d))
\end{equation}

The partial derivative of $f$ with respect to $g_{c\mathbf{i}}^{(r)}$ can be calculated using the chain rule:
\begin{align}
\frac{\partial f}{\partial g_{c\mathbf{i}}^{(r)}}=&\sum\limits_{n=1}^N(\frac{\partial f}{\partial S(\sum\limits_{\mathbf{i}\in\Omega}y^{(n)}_{q\mathbf{i}}g_{c\mathbf{i}},d)}\frac{\partial S(\sum\limits_{\mathbf{i}\in\Omega}y^{(n)}_{q\mathbf{i}}g_{c\mathbf{i}},d)}{\partial g_{c\mathbf{i}}^{(r)}})\\\nonumber
=&\sum\limits_{n=1}^N(p_{q}^{(n)}\frac{\partial S(\sum\limits_{\mathbf{i}\in\Omega}y^{(n)}_{q\mathbf{i}}g_{c\mathbf{i}},d)}{\partial\sum\limits_{\mathbf{i}\in\Omega}y^{(n)}_{q\mathbf{i}}g_{c\mathbf{i}}}\frac{\partial\sum\limits_{\mathbf{i}\in\Omega}y^{(n)}_{q\mathbf{i}}g_{c\mathbf{i}}}{\partial g_{c\mathbf{i}}}\frac{\partial g_{c\mathbf{i}}}{\partial g_{c\mathbf{i}}^{(r)}})\\\nonumber
=&\sum\limits_{n=1}^N(p_{q}^{(n)}y_{q\mathbf{i}}^{(n)}\frac{\partial S(\sum\limits_{\mathbf{i}\in\Omega}y^{(n)}_{q\mathbf{i}}g_{c\mathbf{i}},d)}{\partial\sum\limits_{\mathbf{i}\in\Omega}y^{(n)}_{q\mathbf{i}}g_{c\mathbf{i}}})
\end{align}
and hence from Lemma~\ref{lemma_c} its absolute value $|\frac{\partial f}{\partial g_{c\mathbf{i}}^{(r)}}|$ is upper bounded as follows:
\begin{equation}
|\frac{\partial f}{\partial g_{c\mathbf{i}}^{(r)}}|\leq K\sum\limits_{n=1}^Np_{q}^{(n)}
\end{equation}

We can therefore arrive at the following Lemma regarding learning per-ray errors $\theta^{(q)}$:
\begin{lemma}
\label{lemma_q}
The absolute value of the partial derivative of $\theta^{(q)}$ with respect to $g_{c\mathbf{i}}^{(r)}$ is theoretically upper bounded by the following relation:
\begin{equation}
|\frac{\partial\theta^{(q)}}{\partial g_{c\mathbf{i}}^{(r)}}|\leq(2K\sum\limits_{n=1}^Np_{q}^{(n)})\sqrt{\theta^{(q)}}
\end{equation}
\end{lemma}
\begin{proof}
\begin{align}
|\frac{\partial\theta^{(q)}}{\partial g_{c\mathbf{i}}^{(r)}}|=&2\sqrt{\theta^{(q)}}|\frac{\partial\sqrt{\theta^{(q)}}}{\partial g_{c\mathbf{i}}^{(r)}}|\\\nonumber
=&2\sqrt{\theta^{(q)}}|\frac{\partial}{\partial g_{c\mathbf{i}}^{(r)}}(c_q^{*}-f)|\\\nonumber
=&2\sqrt{\theta^{(q)}}|\frac{\partial f}{\partial g_{c\mathbf{i}}^{(r)}}|
\end{align}

Plugging in (\ref{f_bound}) we have the Lemma.
\end{proof}

\subsection{Limitations on the MSE Loss Gradient}
We define the following appearance grid $\mathcal{G}_{c0}$ and loss value $L_0$:
\begin{definition}
$\mathcal{G}_{c0}$ is the (or ``a") solution to the MSE Loss's optimization problem across appearance grids of VM rank $R_c$, and the MSE loss associated with it is designated $L_0$.
\begin{align*}
\mathcal{G}_{c0}&=\mathop{\arg\min}\limits_{\mathcal{G}_c}L_{\textnormal{MSE}}\\
L_0&=\mathop{\min}\limits_{\mathcal{G}_c}L_{\textnormal{MSE}}\\
\textnormal{s.t. }&\textnormal{rank}(\mathcal{G}_{c})=R_c \nonumber
\end{align*}
\end{definition}

As $\mathcal{G}_{c0}$ is by definition optimal, most of its elements are close to critical points of $L_{\textrm{MSE}}$, and thus we define small positive values $\epsilon^{(r)}_{c\mathbf{i}}$ (which can be treated as constants) such that:
\begin{equation}
|\frac{\partial\theta^{(r)}_0}{\partial g_{c0\mathbf{i}}^{(r)}}|\leq\epsilon^{(r)}_{c\mathbf{i}}
\end{equation}

We then take advantage of the properties of $\mathcal{G}_{c0}$ by first upper bounding $\frac{\partial\theta}{\partial g_{c\mathbf{i}}^{(r)}}-\frac{\partial\theta}{\partial g_{c0\mathbf{i}}^{(r)}}$ as a surrogate for $\frac{\partial\theta}{\partial g_{c\mathbf{i}}^{(r)}}$ using Lemma~\ref{lemma_q}:
\begin{equation}
\label{surrogate}
|\frac{\partial\theta^{(r)}}{\partial g_{c\mathbf{i}}^{(r)}}-\frac{\partial\theta_0^{(r)}}{\partial g_{c0\mathbf{i}}^{(r)}}|\leq 2K\sum\limits_{n=1}^Np_q^{(n)}|\sum\limits_{n=1}^N(p_q^{(n)}(c_q^{(n)}-c_{q0}^{(n)}))|
\end{equation}

Therefore we can prove the following Theorem:
\begin{theorem}
\label{theo}
The iteration step size of gradient descent-based learning on the MSE Loss with respect to each component element of the tensorial appearance grid, $g_{c\mathbf{i}}^{(r)}$, is theoretically limited by the following upper bound on the absolute value of the partial derivative:
\begin{equation}
|\frac{\partial L_{\textnormal{MSE}}}{\partial g_{c\mathbf{i}}^{(r)}}|\leq 2K^2N^2p_{\textnormal{max}}^2(\sum\limits_{\mathbf{i}\in\Omega}|g_{c\mathbf{i}}^{(r)}-g_{c0\mathbf{i}}^{(r)}|)+\epsilon_{c\mathbf{i}}^{(r)}
\end{equation}
where $p_{\textnormal{max}}=\max\limits_{n,q}p_q^{(n)}$.
\end{theorem}
\begin{proof}
From (\ref{surrogate}) we have the following upper bound:
\begin{equation*}
|\frac{\partial L_{\textnormal{MSE}}}{\partial g_{c\mathbf{i}}^{(r)}}|\leq 2K\sum\limits_{n=1}^Np_q^{(n)}|\sum\limits_{n=1}^N(p_q^{(n)}(c_q^{(n)}-c_{q0}^{(n)}))|+\frac{\partial\theta_0^{(r)}}{\partial g_{c0\mathbf{i}}^{(r)}}
\end{equation*}

From (\ref{s_bound}) we trivially have the following:
\begin{equation*}
|c_q^{(n)}-c_{q0}^{(n)}|\leq K\sum\limits_{\mathbf{i}\in\Omega}|g_{c\mathbf{i}}^{(r)}-g_{c0\mathbf{i}}^{(r)}|
\end{equation*}

Plugging in, we have:
\begin{equation*}
|\frac{\partial L_{\textnormal{MSE}}}{\partial g_{c\mathbf{i}}^{(r)}}|\leq 2K^2\sum\limits_{n=1}^Np_q^{(n)}\sum\limits_{n=1}^N(p_q^{(n)}\sum\limits_{\mathbf{i}\in\Omega}|g_c^{(n)}-g_{c0}^{(n)}|)+\epsilon^{(r)}_{c\mathbf{i}}
\end{equation*}

When $p_{\textnormal{max}}=\max\limits_{n,q}p_q^{(n)}$, the above can be loosened and simplified to:
\begin{equation*}
|\frac{\partial L_{\textnormal{MSE}}}{\partial g_{c\mathbf{i}}^{(r)}}|\leq 2K^2N^2p_{\textnormal{max}}^2(\sum\limits_{\mathbf{i}\in\Omega}|g_c^{(r)}-g_{c0}^{(r)}|)+\epsilon^{(r)}_{c\mathbf{i}}
\end{equation*}
\end{proof}

\subsection{Interpretation and Significance}
Theorem~\ref{theo} suggests that the upper bound of $\frac{\partial\theta^{(r)}}{\partial g_{c\mathbf{i}}^{(r)}}$ (intuitively the ``learning speed") is linearly positively correlated with $|g_c^{(r)}-g_{c0}^{(r)}|$ (intuitively the ``distance" between component elements and their ``ideal" values). We explore the implications of this result by investigating two appearance grid components $\mathcal{G}_c^{(r_1)}$ and $\mathcal{G}_c^{(r_2)}$, where $r_1<r_2$.

In our TRaIn algorithm, we initially control $\mathcal{G}_c^{(r_2)}$ when $\mathcal{G}_c^{(r_1)}$ is being trained to keep $|g_c^{(r_2)}-g_{c0}^{(r_2)}|$ constant. This allows for the upper bound from Theorem~\ref{theo} to stay relatively loose. In contrast, most (possibly all) previous tensorial representation-based NeRF paradigms train all components at once, which makes values of $|g_c^{(r)}-g_{c0}^{(r)}|$ for all $r$ lower simultaneously.

Hence, in our algorithm, we allow for $\mathcal{G}_c^{(r_1)}$ to be trained at a speed which cannot be reached by previous paradigms due to the hidden theoretical limit. This is at the expense of $\mathcal{G}_c^{(r_2)}$ being trained slower later due to a higher $|g_c^{(r_1)}-g_{c0}^{(r_1)}|$. Therefore, we achieve slimmability by allowing $\mathcal{G}_c^{(r_1)}$ to capture more information than $\mathcal{G}_c^{(r_2)}$.

\newpage

\begin{table*}[t]
\tabsize
\begin{center}
\begin{tabular}{|c|c|cccccccccccccccc|}
\hline
\multicolumn{2}{|c|}{Rank:} & 1 & 2 & 3 & 4 & 5 & 6 & 7 & 8 & 9 & 10 & 11 & 12 & 13 & 14 & 15 & 16 \\
\hline
\multirow{2}{*}{4}  & Baseline & 15.95 & 17.85 & 20.81 & 34.87 & \multicolumn{12}{|c|}{} \\
                    & SlimmeRF & 16.85 & 29.51 & 31.43 & 34.83 & \multicolumn{12}{|c|}{} \\
                    \hline
\multirow{2}{*}{8}  & Baseline & 14.67 & 15.04 & 17.43 & 19.82 & 21.60 & 24.49 & 27.52 & 35.48 & \multicolumn{8}{|c|}{} \\
                    & SlimmeRF & 19.90 & 27.87 & 30.05 & 32.32 & 33.60 & 35.12 & 35.13 & 35.14 & \multicolumn{8}{|c|}{} \\
                    \hline
\multirow{2}{*}{16} & Baseline & 14.86 & 15.53 & 16.12 & 16.39 & 17.32 & 18.57 & 18.96 & 19.89 & 20.90 & 22.19 & 24.20 & 25.27 & 26.63 & 32.08 & 33.46 & 35.80 \\
                    & SlimmeRF & 16.96 & 23.26 & 28.54 & 30.11 & 33.37 & 35.27 & 35.28 & 35.28 & 35.29 & 35.29 & 35.29 & 35.29 & 35.29 & 35.29 & 35.29 & 35.34 \\
                    \hline
\multirow{5}{*}{32} & Baseline & 14.05 & 14.08 & 14.42 & 14.57 & 14.79 & 14.85 & 15.25 & 15.33 & 16.19 & 16.61 & 18.54 & 18.77 & 19.18 & 20.94 & 21.99 & 22.85 \\
                    & SlimmeRF & 14.97 & 17.48 & 19.08 & 21.28 & 23.30 & 27.22 & 28.20 & 28.96 & 29.66 & 30.05 & 30.52 & 30.97 & 31.44 & 31.84 & 32.14 & 32.43 \\
                    \cline{3-18}
                    & & 17 & 18 & 19 & 20 & 21 & 22 & 23 & 24 & 25 & 26 & 27 & 28 & 29 & 30 & 31 & 32 \\
                    \cline{3-18}
                    & Baseline & 23.89 & 24.71 & 25.06 & 25.58 & 26.11 & 26.87 & 28.51 & 29.20 & 29.58 & 31.29 & 32.27 & 33.86 & 34.77 & 35.13 & 35.45 & 35.95 \\
                    & SlimmeRF & 32.81 & 33.07 & 33.36 & 33.61 & 33.85 & 34.10 & 34.29 & 34.58 & 34.79 & 35.01 & 35.22 & 35.39 & 35.58 & 35.75 & 35.88 & 36.02 \\
                    \hline
\end{tabular}
\end{center}
\caption{Baseline comparison tests on \emph{Chair} of NeRF Synthetic. Numbers are in PSNR (dB).}
\label{chair}
\end{table*}%

\begin{table*}[t]
\tabsize
\begin{center}
\begin{tabular}{|c|c|cccccccccccccccc|}
\hline
\multicolumn{2}{|c|}{Rank:} & 1 & 2 & 3 & 4 & 5 & 6 & 7 & 8 & 9 & 10 & 11 & 12 & 13 & 14 & 15 & 16 \\
\hline
\multirow{2}{*}{4}  & Baseline & 11.67 & 14.00 & 16.57 & 25.58 & \multicolumn{12}{|c|}{} \\
                    & SlimmeRF & 14.09 & 22.34 & 25.25 & 25.40 & \multicolumn{12}{|c|}{} \\
                    \hline
\multirow{2}{*}{8}  & Baseline & 11.45 & 11.93 & 13.11 & 14.66 & 17.39 & 20.33 & 23.49 & 25.77 & \multicolumn{8}{|c|}{} \\
                    & SlimmeRF & 15.33 & 21.86 & 24.14 & 24.89 & 25.44 & 25.55 & 25.60 & 25.63 & \multicolumn{8}{|c|}{} \\
                    \hline
\multirow{2}{*}{16} & Baseline & 11.00 & 11.09 & 11.26 & 11.43 & 12.05 & 12.68 & 13.60 & 14.77 & 15.93 & 16.68 & 18.32 & 19.93 & 22.66 & 24.69 & 25.37 & 26.01 \\
                    & SlimmeRF & 14.51 & 21.19 & 23.00 & 23.93 & 24.71 & 25.05 & 25.28 & 25.45 & 25.51 & 25.55 & 25.56 & 25.57 & 25.58 & 25.58 & 25.58 & 25.58 \\
                    \hline
\multirow{5}{*}{32} & Baseline & 10.96 & 10.96 & 11.30 & 11.34 & 11.43 & 11.52 & 11.59 & 11.69 & 11.86 & 11.95 & 12.49 & 12.68 & 13.01 & 13.73 & 14.43 & 14.98 \\
                    & SlimmeRF & 11.20 & 14.20 & 17.59 & 18.76 & 20.10 & 21.03 & 21.71 & 22.21 & 22.77 & 23.14 & 23.42 & 23.74 & 24.04 & 24.31 & 24.52 & 24.72 \\
                    \cline{3-18}
                    & & 17 & 18 & 19 & 20 & 21 & 22 & 23 & 24 & 25 & 26 & 27 & 28 & 29 & 30 & 31 & 32 \\
                    \cline{3-18}
                    & Baseline & 15.48 & 16.19 & 16.89 & 17.28 & 17.86 & 18.24 & 19.59 & 20.19 & 20.68 & 21.61 & 22.49 & 24.66 & 25.57 & 25.74 & 25.87 & 25.98 \\
                    & SlimmeRF & 24.89 & 25.00 & 25.14 & 25.27 & 25.35 & 25.44 & 25.54 & 25.64 & 25.71 & 25.76 & 25.82 & 25.86 & 25.89 & 25.93 & 25.96 & 25.98 \\
                    \hline
\end{tabular}
\end{center}
\caption{Baseline comparison tests on \emph{Drums} of NeRF Synthetic. Numbers are in PSNR (dB).}
\label{drums}
\end{table*}%

\begin{table*}[t]
\tabsize
\begin{center}
\begin{tabular}{|c|c|cccccccccccccccc|}
\hline
\multicolumn{2}{|c|}{Rank:} & 1 & 2 & 3 & 4 & 5 & 6 & 7 & 8 & 9 & 10 & 11 & 12 & 13 & 14 & 15 & 16 \\
\hline
\multirow{2}{*}{4}  & Baseline & 15.80 & 18.23 & 22.73 & 32.63 & \multicolumn{12}{|c|}{} \\
                    & SlimmeRF & 16.31 & 20.15 & 24.32 & 32.63 & \multicolumn{12}{|c|}{} \\
                    \hline
\multirow{2}{*}{8}  & Baseline & 14.34 & 14.71 & 15.68 & 17.26 & 19.44 & 23.69 & 28.25 & 33.69 & \multicolumn{8}{|c|}{} \\
                    & SlimmeRF & 14.41 & 16.32 & 17.28 & 18.89 & 21.22 & 24.31 & 28.50 & 33.74 & \multicolumn{8}{|c|}{} \\
                    \hline
\multirow{2}{*}{16} & Baseline & 14.24 & 14.26 & 14.36 & 14.56 & 14.93 & 15.54 & 16.51 & 17.80 & 19.50 & 21.03 & 22.67 & 24.69 & 27.18 & 29.57 & 31.85 & 34.11 \\
                    & SlimmeRF & 14.60 & 16.65 & 17.13 & 17.76 & 18.51 & 19.40 & 20.43 & 21.60 & 23.04 & 24.46 & 26.41 & 28.14 & 29.77 & 31.48 & 33.03 & 34.03 \\
                    \hline
\multirow{5}{*}{32} & Baseline & 14.24 & 14.24 & 14.27 & 14.29 & 14.35 & 14.41 & 14.47 & 14.53 & 14.61 & 14.74 & 15.44 & 15.72 & 15.94 & 16.46 & 17.10 & 17.91 \\
                    & SlimmeRF & 14.93 & 15.89 & 16.69 & 17.21 & 17.73 & 18.37 & 19.06 & 20.05 & 20.87 & 21.81 & 22.65 & 23.57 & 24.47 & 25.10 & 25.96 & 26.72 \\
                    \cline{3-18}
                    & & 17 & 18 & 19 & 20 & 21 & 22 & 23 & 24 & 25 & 26 & 27 & 28 & 29 & 30 & 31 & 32 \\
                    \cline{3-18}
                    & Baseline & 18.75 & 19.65 & 20.29 & 20.80 & 21.43 & 22.28 & 23.03 & 24.81 & 26.15 & 28.00 & 29.29 & 30.29 & 32.07 & 32.75 & 33.60 & 34.14 \\
                    & SlimmeRF & 27.49 & 28.25 & 28.97 & 29.70 & 30.32 & 30.82 & 31.32 & 31.82 & 32.16 & 32.57 & 32.97 & 33.34 & 33.66 & 33.83 & 34.04 & 34.17 \\
                    \hline
\end{tabular}
\end{center}
\caption{Baseline comparison tests on \emph{Ficus} of NeRF Synthetic. Numbers are in PSNR (dB).}
\label{ficus}
\end{table*}%

\begin{table*}[t]
\tabsize
\begin{center}
\begin{tabular}{|c|c|cccccccccccccccc|}
\hline
\multicolumn{2}{|c|}{Rank:} & 1 & 2 & 3 & 4 & 5 & 6 & 7 & 8 & 9 & 10 & 11 & 12 & 13 & 14 & 15 & 16 \\
\hline
\multirow{2}{*}{4}  & Baseline & 12.19 & 15.13 & 20.01 & 36.81 & \multicolumn{12}{|c|}{} \\
                    & SlimmeRF & 14.87 & 30.17 & 31.87 & 36.70 & \multicolumn{12}{|c|}{} \\
                    \hline
\multirow{2}{*}{8}  & Baseline & 11.06 & 12.52 & 13.47 & 14.66 & 15.42 & 24.43 & 30.83 & 37.22 & \multicolumn{8}{|c|}{} \\
                    & SlimmeRF & 15.55 & 29.32 & 32.83 & 33.98 & 34.95 & 35.78 & 36.40 & 37.01 & \multicolumn{8}{|c|}{} \\
                    \hline
\multirow{2}{*}{16} & Baseline & 10.45 & 10.47 & 10.48 & 10.62 & 10.88 & 11.06 & 12.36 & 13.74 & 14.45 & 15.13 & 16.06 & 16.93 & 20.49 & 26.02 & 29.29 & 37.57 \\
                    & SlimmeRF & 15.89 & 28.54 & 31.66 & 32.90 & 33.62 & 34.07 & 35.03 & 35.55 & 35.91 & 36.12 & 36.43 & 36.63 & 36.80 & 36.96 & 37.10 & 37.25 \\
                    \hline
\multirow{5}{*}{32} & Baseline & 10.40 & 10.58 & 10.65 & 10.66 & 10.80 & 11.06 & 11.13 & 11.26 & 11.37 & 11.51 & 11.92 & 12.01 & 12.16 & 12.28 & 12.98 & 17.61 \\
                    & SlimmeRF & 14.59 & 27.18 & 30.56 & 32.79 & 34.00 & 34.69 & 35.04 & 35.48 & 35.74 & 35.98 & 36.18 & 36.33 & 36.46 & 36.57 & 36.66 & 36.75 \\
                    \cline{3-18}
                    & & 17 & 18 & 19 & 20 & 21 & 22 & 23 & 24 & 25 & 26 & 27 & 28 & 29 & 30 & 31 & 32 \\
                    \cline{3-18}
                    & Baseline & 18.17 & 18.39 & 18.76 & 18.94 & 19.68 & 22.23 & 22.83 & 24.48 & 25.01 & 28.26 & 28.54 & 32.70 & 34.00 & 34.31 & 35.22 & 37.66 \\
                    & SlimmeRF & 36.85 & 36.90 & 36.98 & 37.02 & 37.07 & 37.13 & 37.18 & 37.21 & 37.25 & 37.29 & 37.33 & 37.35 & 37.38 & 37.39 & 37.42 & 37.43 \\
                    \hline
\end{tabular}
\end{center}
\caption{Baseline comparison tests on \emph{Hotdog} of NeRF Synthetic. Numbers are in PSNR (dB).}
\label{hotdog}
\end{table*}%

\begin{table*}[t]
\tabsize
\begin{center}
\begin{tabular}{|c|c|cccccccccccccccc|}
\hline
\multicolumn{2}{|c|}{Rank:} & 1 & 2 & 3 & 4 & 5 & 6 & 7 & 8 & 9 & 10 & 11 & 12 & 13 & 14 & 15 & 16 \\
\hline
\multirow{2}{*}{4}  & Baseline & 11.47 & 14.12 & 19.65 & 35.47 & \multicolumn{12}{|c|}{} \\
                    & SlimmeRF & 13.15 & 20.86 & 25.35 & 35.63 & \multicolumn{12}{|c|}{} \\
                    \hline
\multirow{2}{*}{8}  & Baseline & 10.23 & 11.16 & 12.74 & 15.68 & 19.14 & 23.78 & 29.25 & 36.18 & \multicolumn{8}{|c|}{} \\
                    & SlimmeRF & 14.77 & 23.66 & 26.64 & 28.66 & 32.39 & 33.61 & 34.95 & 36.11 & \multicolumn{8}{|c|}{} \\
                    \hline
\multirow{2}{*}{16} & Baseline &  9.65 & 10.00 & 10.49 & 11.31 & 12.90 & 14.72 & 15.43 & 16.76 & 17.89 & 19.99 & 21.45 & 22.42 & 25.53 & 28.16 & 31.98 & 36.54 \\
                    & SlimmeRF & 13.14 & 21.40 & 24.90 & 27.72 & 29.86 & 31.46 & 32.86 & 33.53 & 34.02 & 34.60 & 34.96 & 35.43 & 35.82 & 36.04 & 36.26 & 36.45 \\
                    \hline
\multirow{5}{*}{32} & Baseline &  9.48 &  9.49 &  9.53 &  9.54 &  9.57 &  9.63 &  9.74 &  9.81 &  9.94 & 10.21 & 10.71 & 10.86 & 10.99 & 11.81 & 12.16 & 12.32 \\
                    & SlimmeRF & 13.75 & 23.41 & 25.22 & 27.59 & 29.47 & 30.72 & 31.57 & 32.74 & 33.31 & 33.69 & 34.03 & 34.29 & 34.81 & 35.10 & 35.26 & 35.47 \\
                    \cline{3-18}
                    & & 17 & 18 & 19 & 20 & 21 & 22 & 23 & 24 & 25 & 26 & 27 & 28 & 29 & 30 & 31 & 32 \\
                    \cline{3-18}
                    & Baseline & 13.14 & 13.98 & 14.86 & 15.97 & 16.81 & 17.14 & 19.49 & 22.23 & 24.09 & 26.66 & 27.54 & 31.00 & 33.35 & 35.22 & 35.99 & 36.79 \\
                    & SlimmeRF & 35.59 & 35.71 & 35.84 & 35.92 & 36.03 & 36.07 & 36.17 & 36.22 & 36.25 & 36.30 & 36.33 & 36.35 & 36.37 & 36.39 & 36.41 & 36.43 \\
                    \hline
\end{tabular}
\end{center}
\caption{Baseline comparison tests on \emph{Lego} of NeRF Synthetic. Numbers are in PSNR (dB).}
\label{lego}
\end{table*}%

\begin{table*}[t]
\tabsize
\begin{center}
\begin{tabular}{|c|c|cccccccccccccccc|}
\hline
\multicolumn{2}{|c|}{Rank:} & 1 & 2 & 3 & 4 & 5 & 6 & 7 & 8 & 9 & 10 & 11 & 12 & 13 & 14 & 15 & 16 \\
\hline
\multirow{2}{*}{4}  & Baseline & 10.07 & 12.86 & 19.11 & 29.54 & \multicolumn{12}{|c|}{} \\
                    & SlimmeRF & 13.27 & 20.48 & 24.78 & 29.50 & \multicolumn{12}{|c|}{} \\
                    \hline
\multirow{2}{*}{8}  & Baseline &  8.87 &  9.68 & 12.34 & 14.73 & 18.42 & 23.23 & 26.39 & 29.90 & \multicolumn{8}{|c|}{} \\
                    & SlimmeRF & 11.20 & 19.81 & 23.55 & 25.50 & 26.52 & 27.71 & 28.88 & 29.86 & \multicolumn{8}{|c|}{} \\
                    \hline
\multirow{2}{*}{16} & Baseline &  8.74 &  8.74 &  8.75 &  8.76 &  8.84 &  9.14 & 10.36 & 15.41 & 16.18 & 16.36 & 20.38 & 21.70 & 23.15 & 24.79 & 28.54 & 30.09 \\
                    & SlimmeRF & 12.40 & 20.00 & 23.34 & 24.99 & 25.67 & 26.26 & 26.97 & 27.76 & 28.52 & 28.90 & 29.16 & 29.42 & 29.59 & 29.77 & 29.93 & 30.04 \\
                    \hline
\multirow{5}{*}{32} & Baseline &  8.74 &  8.74 &  9.06 &  9.07 &  9.24 &  9.39 &  9.48 &  9.53 &  9.61 &  9.64 &  9.79 &  9.93 & 10.01 & 12.74 & 12.81 & 13.16 \\
                    & SlimmeRF & 11.62 & 19.58 & 22.53 & 24.53 & 25.78 & 26.32 & 26.92 & 27.32 & 27.92 & 28.31 & 28.61 & 28.89 & 29.05 & 29.14 & 29.27 & 29.36 \\
                    \cline{3-18}
                    & & 17 & 18 & 19 & 20 & 21 & 22 & 23 & 24 & 25 & 26 & 27 & 28 & 29 & 30 & 31 & 32 \\
                    \cline{3-18}
                    & Baseline & 14.21 & 15.42 & 17.02 & 17.15 & 17.75 & 18.40 & 19.15 & 19.49 & 19.57 & 25.72 & 27.75 & 28.08 & 28.50 & 28.70 & 29.34 & 30.19 \\
                    & SlimmeRF & 29.49 & 29.57 & 29.65 & 29.72 & 29.79 & 29.83 & 29.89 & 29.95 & 30.00 & 30.03 & 30.07 & 30.09 & 30.12 & 30.14 & 30.16 & 30.18 \\
                    \hline
\end{tabular}
\end{center}
\caption{Baseline comparison tests on \emph{Materials} of NeRF Synthetic. Numbers are in PSNR (dB).}
\label{materials}
\end{table*}%

\begin{table*}[t]
\tabsize
\begin{center}
\begin{tabular}{|c|c|cccccccccccccccc|}
\hline
\multicolumn{2}{|c|}{Rank:} & 1 & 2 & 3 & 4 & 5 & 6 & 7 & 8 & 9 & 10 & 11 & 12 & 13 & 14 & 15 & 16 \\
\hline
\multirow{2}{*}{4}  & Baseline & 14.33 & 16.96 & 22.35 & 33.85 & \multicolumn{12}{|c|}{} \\
                    & SlimmeRF & 15.58 & 21.75 & 29.30 & 34.20 & \multicolumn{12}{|c|}{} \\
                    \hline
\multirow{2}{*}{8}  & Baseline & 13.30 & 13.53 & 16.18 & 18.60 & 20.69 & 23.49 & 25.31 & 34.44 & \multicolumn{8}{|c|}{} \\
                    & SlimmeRF & 16.29 & 24.01 & 27.04 & 28.22 & 29.70 & 31.37 & 32.81 & 34.51 & \multicolumn{8}{|c|}{} \\
                    \hline
\multirow{2}{*}{16} & Baseline & 13.21 & 14.10 & 14.87 & 15.42 & 15.82 & 16.34 & 17.40 & 18.28 & 19.97 & 22.16 & 23.15 & 24.07 & 27.37 & 31.72 & 32.89 & 34.97 \\
                    & SlimmeRF & 17.27 & 23.80 & 26.33 & 27.58 & 28.28 & 29.27 & 30.08 & 30.80 & 31.60 & 31.88 & 32.45 & 32.88 & 33.25 & 33.75 & 34.35 & 35.06 \\
                    \hline
\multirow{5}{*}{32} & Baseline & 13.04 & 13.16 & 13.97 & 14.07 & 14.10 & 14.14 & 14.20 & 14.33 & 14.60 & 14.73 & 15.64 & 15.74 & 15.90 & 16.72 & 19.56 & 20.34 \\
                    & SlimmeRF & 14.55 & 24.11 & 26.81 & 28.09 & 28.47 & 29.18 & 29.93 & 30.49 & 30.79 & 31.00 & 31.33 & 31.47 & 31.78 & 31.93 & 32.07 & 32.26 \\
                    \cline{3-18}
                    & & 17 & 18 & 19 & 20 & 21 & 22 & 23 & 24 & 25 & 26 & 27 & 28 & 29 & 30 & 31 & 32 \\
                    \cline{3-18}
                    & Baseline & 20.68 & 20.91 & 22.81 & 23.02 & 23.31 & 23.85 & 25.12 & 25.87 & 26.55 & 29.08 & 29.46 & 30.36 & 32.29 & 32.85 & 33.87 & 34.99 \\
                    & SlimmeRF & 32.58 & 32.75 & 32.93 & 33.13 & 33.23 & 33.37 & 33.50 & 33.87 & 34.03 & 34.17 & 34.28 & 34.50 & 34.75 & 34.90 & 34.95 & 35.00 \\
                    \hline
\end{tabular}
\end{center}
\caption{Baseline comparison tests on \emph{Mic} of NeRF Synthetic. Numbers are in PSNR (dB).}
\label{mic}
\end{table*}%

\begin{table*}[t]
\tabsize
\begin{center}
\begin{tabular}{|c|c|cccccccccccccccc|}
\hline
\multicolumn{2}{|c|}{Rank:} & 1 & 2 & 3 & 4 & 5 & 6 & 7 & 8 & 9 & 10 & 11 & 12 & 13 & 14 & 15 & 16 \\
\hline
\multirow{2}{*}{4}  & Baseline &  7.62 & 11.25 & 18.60 & 29.95 & \multicolumn{12}{|c|}{} \\
                    & SlimmeRF & 11.98 & 23.58 & 26.61 & 29.91 & \multicolumn{12}{|c|}{} \\
                    \hline
\multirow{2}{*}{8}  & Baseline &  6.15 &  6.94 & 10.72 & 13.53 & 14.88 & 16.89 & 23.43 & 30.46 & \multicolumn{8}{|c|}{} \\
                    & SlimmeRF & 10.75 & 21.81 & 24.62 & 26.26 & 27.70 & 28.71 & 29.69 & 30.43 & \multicolumn{8}{|c|}{} \\
                    \hline
\multirow{2}{*}{16} & Baseline &  6.03 &  6.11 &  6.23 &  6.86 &  7.65 &  8.50 & 13.26 & 14.81 & 16.32 & 18.24 & 20.05 & 22.95 & 24.98 & 27.65 & 28.81 & 30.72 \\
                    & SlimmeRF & 11.85 & 21.98 & 23.99 & 25.41 & 26.50 & 27.36 & 28.11 & 28.69 & 29.11 & 29.59 & 29.91 & 30.18 & 30.39 & 30.52 & 30.56 & 30.58 \\
                    \hline
\multirow{5}{*}{32} & Baseline &  5.90 &  5.97 &  6.22 &  6.27 &  6.30 &  6.87 &  7.00 &  7.20 &  7.42 &  7.51 &  7.90 &  7.98 &  8.24 &  8.61 &  9.08 &  9.31 \\
                    & SlimmeRF & 10.07 & 22.57 & 24.01 & 25.00 & 26.29 & 27.23 & 27.76 & 28.34 & 28.71 & 29.17 & 29.51 & 29.87 & 30.16 & 30.38 & 30.54 & 30.55 \\
                    \cline{3-18}
                    & & 17 & 18 & 19 & 20 & 21 & 22 & 23 & 24 & 25 & 26 & 27 & 28 & 29 & 30 & 31 & 32 \\
                    \cline{3-18}
                    & Baseline &  9.45 & 10.52 & 10.68 & 11.37 & 12.21 & 12.64 & 13.23 & 25.92 & 26.48 & 27.34 & 27.80 & 28.75 & 29.18 & 29.66 & 30.46 & 30.86 \\
                    & SlimmeRF & 30.56 & 30.56 & 30.56 & 30.56 & 30.56 & 30.56 & 30.56 & 30.56 & 30.56 & 30.56 & 30.56 & 30.56 & 30.56 & 30.56 & 30.56 & 30.56 \\
                    \hline
\end{tabular}
\end{center}
\caption{Baseline comparison tests on \emph{Ship} of NeRF Synthetic. Numbers are in PSNR (dB).}
\label{ship}
\end{table*}%

\begin{table*}[t]
\tabsize
\begin{center}
\begin{tabular}{|c|c|cccccccccccccccc|}
\hline
\multicolumn{2}{|c|}{Rank:} & 1 & 2 & 3 & 4 & 5 & 6 & 7 & 8 & 9 & 10 & 11 & 12 & 13 & 14 & 15 & 16 \\
\hline
\multirow{2}{*}{4}  & Baseline & 12.39 & 15.05 & 19.98 & 32.34 & \multicolumn{12}{|c|}{} \\
                    & SlimmeRF & 14.51 & 23.60 & 27.36 & 32.35 & \multicolumn{12}{|c|}{} \\
                    \hline
\multirow{2}{*}{8}  & Baseline & 11.26 & 11.94 & 13.96 & 16.12 & 18.37 & 22.54 & 26.81 & 32.89 & \multicolumn{8}{|c|}{} \\
                    & SlimmeRF & 14.78 & 23.08 & 25.77 & 27.34 & 28.94 & 30.27 & 31.50 & 32.80 & \multicolumn{8}{|c|}{} \\
                    \hline
\multirow{2}{*}{16} & Baseline & 11.02 & 11.29 & 11.57 & 11.92 & 12.55 & 13.32 & 14.74 & 16.43 & 17.64 & 18.97 & 20.78 & 22.25 & 24.75 & 28.09 & 30.27 & 33.23 \\
                    & SlimmeRF & 14.58 & 22.10 & 24.86 & 26.30 & 27.56 & 28.52 & 29.26 & 29.83 & 30.37 & 30.80 & 31.27 & 31.69 & 32.06 & 32.42 & 32.76 & 33.04 \\
                    \hline
\multirow{5}{*}{32} & Baseline & 10.85 & 10.90 & 11.18 & 11.23 & 11.32 & 11.48 & 11.61 & 11.71 & 11.95 & 12.11 & 12.80 & 12.96 & 13.18 & 14.16 & 15.01 & 16.06 \\
                    & SlimmeRF & 13.21 & 20.55 & 22.81 & 24.41 & 25.64 & 26.84 & 27.53 & 28.20 & 28.72 & 29.14 & 29.53 & 29.89 & 30.28 & 30.55 & 30.80 & 31.03 \\
                    \cline{3-18}
                    & & 17 & 18 & 19 & 20 & 21 & 22 & 23 & 24 & 25 & 26 & 27 & 28 & 29 & 30 & 31 & 32 \\
                    \cline{3-18}
                    & Baseline & 16.72 & 17.47 & 18.30 & 18.77 & 19.39 & 20.21 & 21.37 & 24.02 & 24.76 & 27.24 & 28.14 & 29.96 & 31.22 & 31.80 & 32.48 & 33.32 \\
                    & SlimmeRF & 31.28 & 31.48 & 31.68 & 31.87 & 32.02 & 32.17 & 32.31 & 32.48 & 32.59 & 32.71 & 32.82 & 32.93 & 33.04 & 33.11 & 33.17 & 33.22 \\
                    \hline
\end{tabular}
\end{center}
\caption{Average results of baseline comparison tests on NeRF Synthetic. Numbers are in PSNR (dB).}
\label{synth}
\end{table*}%

\begin{table*}[t]
\ssmall
\centering
\begin{tabular}{|c|c|cccccccccccccccc|}
\hline
\multicolumn{2}{|c|}{Rank:} & 1 & 2 & 3 & 4 & 5 & 6 & 7 & 8 & 9 & 10 & 11 & 12 & 13 & 14 & 15 & 16 \\
\hline
\multirow{2}{*}{\emph{Barn}}  & Baseline & 8.19 & 8.63 & 9.28 & 9.90 & 10.75 & 11.43 & 11.90 & 14.91 & 15.96 & 17.17 & 18.04 & 18.95 & 22.22 & 23.88 & 25.30 & 27.44 \\
                    & SlimmeRF & 12.24 & 15.87 & 19.69 & 21.64 & 22.28 & 23.10 & 23.65 & 24.34 & 24.76 & 25.23 & 25.64 & 26.03 & 26.49 & 26.76 & 27.01 & 27.18 \\
                    \hline
\multirow{2}{*}{\emph{Caterpillar}}  & Baseline & 8.20 & 8.48 & 9.48 & 9.80 & 10.24 & 12.20 & 15.40 & 15.86 & 17.49 & 19.02 & 20.00 & 21.33 & 22.60 & 23.47 & 25.00 & 26.00 \\
                    & SlimmeRF & 10.46 & 16.59 & 18.53 & 19.78 & 21.11 & 21.85 & 22.62 & 23.05 & 23.55 & 24.10 & 24.50 & 24.81 & 25.08 & 25.42 & 25.66 & 25.98 \\
                    \hline
\multirow{2}{*}{\emph{Family}} & Baseline & 11.69 & 12.62 & 14.26 & 15.64 & 17.02 & 20.33 & 21.91 & 24.37 & 25.43 & 26.54 & 27.84 & 29.34 & 30.66 & 31.67 & 32.81 & 34.10 \\
                    & SlimmeRF & 18.11 & 22.56 & 24.25 & 25.19 & 26.12 & 26.97 & 27.91 & 28.63 & 29.50 & 30.22 & 30.88 & 31.76 & 32.27 & 32.74 & 33.34 & 34.03 \\
                    \hline
\multirow{2}{*}{\emph{Ignatius}} & Baseline & 12.89 & 12.91 & 13.03 & 13.30 & 13.66 & 13.82 & 14.78 & 16.25 & 19.10 & 20.80 & 21.88 & 23.19 & 25.37 & 26.63 & 27.34 & 28.49 \\
                    & SlimmeRF & 20.66 & 22.82 & 24.51 & 25.13 & 25.65 & 26.34 & 26.70 & 27.04 & 27.30 & 27.44 & 27.64 & 27.88 & 28.04 & 28.17 & 28.38 & 28.46 \\
                    \hline
\multirow{2}{*}{\emph{Truck}} & Baseline & 9.20 & 9.51 & 9.87 & 10.66 & 11.31 & 11.65 & 12.72 & 14.09 & 15.90 & 17.73 & 18.85 & 21.06 & 22.81 & 24.02 & 25.33 & 26.85 \\
                    & SlimmeRF & 12.61 & 14.87 & 17.28 & 19.89 & 21.16 & 22.10 & 22.92 & 23.47 & 24.46 & 24.94 & 25.39 & 25.78 & 26.09 & 26.46 & 26.71 & 26.94 \\
                    \hline
\multirow{2}{*}{\emph{Average}} & Baseline & 10.03 & 10.43 & 11.19 & 11.86 & 12.59 & 13.89 & 15.34 & 17.10 & 18.78 & 20.25 & 21.32 & 22.77 & 24.73 & 25.94 & 27.16 & 28.58 \\
                    & SlimmeRF & 14.82 & 18.54 & 20.85 & 22.33 & 23.26 & 24.07 & 24.76 & 25.31 & 25.91 & 26.38 & 26.81 & 27.25 & 27.59 & 27.91 & 28.22 & 28.52 \\
                    \hline
\end{tabular}
\caption{Results of SlimmeRF-16 and the TensoRF-VM-192 Baseline on the Tanks \& Temples Dataset. Numbers are in PSNR (dB).}
\label{tnt}
\end{table*}%

\begin{table*}
\tabsize
\centering
\begin{tabular}{|c|cccccccc|}
\hline
3 Views & Fern & Flower & Fortress & Horns & Leaves & Orchids & Room & T-Rex \\
\hline
1  & 14.68 & 14.37 & 15.13 & 13.66 & 12.29 & 12.13 & 13.85 & 13.88 \\
2  & 16.56 & 15.43 & 17.39 & 16.07 & 13.73 & 13.31 & 16.10 & 15.52 \\
3  & 17.11 & 16.96 & 17.46 & 17.03 & 14.37 & 13.60 & 17.37 & 16.23 \\
4  & 17.20 & 18.14 & 17.47 & 17.53 & 14.73 & 13.82 & 17.44 & 16.69 \\
5  & 17.19 & 18.51 & 17.42 & 17.67 & 14.77 & 13.86 & 17.44 & 16.77 \\
6  & 17.24 & 18.58 & 17.36 & 17.69 & 14.89 & 13.87 & 17.44 & 16.78 \\
7  & 17.23 & 18.62 & 17.33 & 17.76 & 14.93 & 13.88 & 17.43 & 16.80 \\
8  & 17.24 & 18.63 & 17.33 & 17.77 & 14.93 & 13.90 & 17.42 & 16.82 \\
9  & 17.24 & 18.63 & 17.30 & 17.79 & 14.94 & 13.93 & 17.42 & 16.83 \\
10 & 17.23 & 18.63 & 17.28 & 17.80 & 14.95 & 13.92 & 17.41 & 16.83 \\
11 & 17.22 & 18.63 & 17.27 & 17.80 & 14.95 & 13.91 & 17.41 & 16.82 \\
12 & 17.22 & 18.63 & 17.26 & 17.80 & 14.95 & 13.90 & 17.41 & 16.84 \\
13 & 17.21 & 18.63 & 17.26 & 17.80 & 14.94 & 13.93 & 17.41 & 16.84 \\
14 & 17.20 & 18.63 & 17.25 & 17.82 & 14.93 & 13.93 & 17.41 & 16.84 \\
15 & 17.20 & 18.64 & 17.25 & 17.82 & 14.92 & 13.93 & 17.41 & 16.83 \\
16 & 17.19 & 18.63 & 17.24 & 17.82 & 14.91 & 13.93 & 17.40 & 16.83 \\
17 & 17.18 & 18.63 & 17.24 & 17.80 & 14.91 & 13.93 & 17.40 & 16.82 \\
18 & 17.17 & 18.63 & 17.23 & 17.81 & 14.90 & 13.93 & 17.40 & 16.82 \\
19 & 17.16 & 18.63 & 17.23 & 17.81 & 14.89 & 13.93 & 17.40 & 16.82 \\
20 & 17.15 & 18.63 & 17.23 & 17.81 & 14.88 & 13.92 & 17.40 & 16.82 \\
21 & 17.14 & 18.63 & 17.23 & 17.82 & 14.88 & 13.92 & 17.40 & 16.82 \\
22 & 17.13 & 18.63 & 17.22 & 17.82 & 14.87 & 13.92 & 17.39 & 16.81 \\
23 & 17.12 & 18.63 & 17.22 & 17.81 & 14.87 & 13.92 & 17.39 & 16.81 \\
24 & 17.12 & 18.63 & 17.22 & 17.82 & 14.86 & 13.92 & 17.38 & 16.81 \\
\hline
\end{tabular}
\captionsetup{width=.8\textwidth}
\caption{Per-scene results of the experiments with LLFF (3 Views). The leftmost column displays the number of components left. The model used was SlimmeRF-24. Numbers are in PSNR (dB).}
\label{LLFF1}
\end{table*}

\begin{table*}
\tabsize
\centering
\begin{tabular}{|c|cccccccc|}
\hline
6 Views & Fern & Flower & Fortress & Horns & Leaves & Orchids & Room & T-Rex \\
\hline
1  & 16.03 & 15.47 & 16.81 & 13.76 & 12.84 & 12.90 & 14.76 & 14.46 \\
2  & 18.27 & 16.39 & 19.19 & 15.67 & 14.47 & 13.74 & 17.25 & 15.25 \\
3  & 19.50 & 17.80 & 20.14 & 17.13 & 15.09 & 14.49 & 19.45 & 17.41 \\
4  & 20.00 & 18.92 & 20.29 & 17.87 & 15.81 & 15.01 & 20.06 & 17.75 \\
5  & 20.41 & 19.38 & 20.41 & 18.04 & 16.15 & 15.44 & 20.70 & 17.89 \\
6  & 20.66 & 19.76 & 20.48 & 18.20 & 16.46 & 15.64 & 20.86 & 18.01 \\
7  & 20.78 & 19.92 & 20.53 & 18.30 & 16.63 & 15.81 & 20.90 & 18.21 \\
8  & 20.88 & 19.99 & 20.67 & 18.34 & 16.76 & 15.95 & 20.90 & 18.26 \\
9  & 21.00 & 20.00 & 20.63 & 18.39 & 16.77 & 16.15 & 20.91 & 18.31 \\
10 & 21.02 & 20.01 & 20.65 & 18.40 & 16.81 & 16.19 & 20.97 & 18.35 \\
11 & 21.05 & 20.03 & 20.64 & 18.40 & 16.82 & 16.22 & 21.00 & 18.37 \\
12 & 21.07 & 20.05 & 20.64 & 18.40 & 16.81 & 16.24 & 21.02 & 18.37 \\
13 & 21.08 & 20.06 & 20.62 & 18.40 & 16.81 & 16.26 & 21.03 & 18.38 \\
14 & 21.08 & 20.06 & 20.62 & 18.42 & 16.81 & 16.28 & 21.03 & 18.42 \\
15 & 21.08 & 20.06 & 20.61 & 18.45 & 16.80 & 16.28 & 21.03 & 18.42 \\
16 & 21.07 & 20.07 & 20.61 & 18.45 & 16.80 & 16.28 & 21.03 & 18.42 \\
17 & 21.07 & 20.08 & 20.61 & 18.45 & 16.79 & 16.28 & 21.03 & 18.42 \\
18 & 21.07 & 20.08 & 20.60 & 18.45 & 16.79 & 16.29 & 21.02 & 18.42 \\
19 & 21.07 & 20.09 & 20.59 & 18.45 & 16.78 & 16.29 & 21.02 & 18.42 \\
20 & 21.07 & 20.09 & 20.57 & 18.46 & 16.78 & 16.29 & 21.02 & 18.42 \\
21 & 21.06 & 20.09 & 20.56 & 18.46 & 16.77 & 16.29 & 21.01 & 18.42 \\
22 & 21.06 & 20.09 & 20.55 & 18.46 & 16.77 & 16.30 & 21.00 & 18.41 \\
23 & 21.05 & 20.09 & 20.54 & 18.48 & 16.77 & 16.29 & 21.00 & 18.41 \\
24 & 21.05 & 20.09 & 20.53 & 18.47 & 16.76 & 16.29 & 21.00 & 18.41 \\
\hline
\end{tabular}
\captionsetup{width=.8\textwidth}
\caption{Per-scene results of the experiments with LLFF (6 Views). The leftmost column displays the number of components left. The model used was SlimmeRF-24. Numbers are in PSNR (dB).}
\label{LLFF2}
\end{table*}

\begin{table*}
\tabsize
\centering
\begin{tabular}{|c|cccccccc|}
\hline
9 Views & Fern & Flower & Fortress & Horns & Leaves & Orchids & Room & T-Rex \\
\hline
1  & 14.36 & 15.95 & 14.35 & 14.04 & 13.50 & 13.36 & 13.97 & 14.86 \\
2  & 17.24 & 18.19 & 20.64 & 16.35 & 14.71 & 14.05 & 19.19 & 16.32 \\
3  & 20.10 & 19.68 & 21.32 & 17.48 & 15.84 & 15.38 & 22.44 & 18.19 \\
4  & 21.26 & 20.51 & 21.81 & 18.20 & 16.66 & 16.34 & 23.76 & 18.73 \\
5  & 21.73 & 21.12 & 21.99 & 18.82 & 17.01 & 17.04 & 24.13 & 19.15 \\
6  & 22.17 & 21.90 & 22.15 & 19.11 & 17.38 & 17.39 & 24.41 & 19.59 \\
7  & 22.63 & 22.07 & 22.34 & 19.38 & 17.78 & 17.72 & 24.55 & 20.02 \\
8  & 23.00 & 22.19 & 22.41 & 19.55 & 17.95 & 17.91 & 24.64 & 20.13 \\
9  & 23.21 & 22.21 & 22.42 & 19.66 & 18.04 & 18.01 & 24.73 & 20.22 \\
10 & 23.34 & 22.23 & 22.48 & 19.72 & 18.08 & 18.04 & 24.83 & 20.25 \\
11 & 23.44 & 22.24 & 22.49 & 19.73 & 18.12 & 18.08 & 24.85 & 20.34 \\
12 & 23.62 & 22.26 & 22.50 & 19.77 & 18.12 & 18.11 & 24.94 & 20.38 \\
13 & 23.72 & 22.26 & 22.51 & 19.79 & 18.16 & 18.13 & 24.99 & 20.40 \\
14 & 23.77 & 22.26 & 22.51 & 19.80 & 18.17 & 18.13 & 25.02 & 20.43 \\
15 & 23.86 & 22.26 & 22.51 & 19.83 & 18.15 & 18.13 & 25.05 & 20.43 \\
16 & 23.93 & 22.27 & 22.52 & 19.82 & 18.15 & 18.13 & 25.09 & 20.45 \\
17 & 24.02 & 22.27 & 22.52 & 19.82 & 18.14 & 18.13 & 25.11 & 20.48 \\
18 & 24.04 & 22.27 & 22.52 & 19.84 & 18.14 & 18.13 & 25.13 & 20.48 \\
19 & 24.06 & 22.27 & 22.52 & 19.84 & 18.14 & 18.14 & 25.17 & 20.50 \\
20 & 24.08 & 22.27 & 22.53 & 19.86 & 18.14 & 18.14 & 25.18 & 20.51 \\
21 & 24.08 & 22.27 & 22.53 & 19.85 & 18.13 & 18.13 & 25.18 & 20.52 \\
22 & 24.11 & 22.27 & 22.53 & 19.85 & 18.13 & 18.13 & 25.18 & 20.51 \\
23 & 24.14 & 22.27 & 22.53 & 19.87 & 18.13 & 18.13 & 25.18 & 20.51 \\
24 & 24.15 & 22.27 & 22.53 & 19.87 & 18.12 & 18.13 & 25.18 & 20.51 \\
\hline
\end{tabular}
\captionsetup{width=.8\textwidth}
\caption{Per-scene results of the experiments with LLFF (9 Views). The leftmost column displays the number of components left. The model used was SlimmeRF-24. Numbers are in PSNR (dB).}
\label{LLFF3}
\end{table*}

\begin{table*}
\tabsize
\centering
\begin{tabular}{|c|c|c|c|c|c|c|c|c|c|c|c|c|}
\hline
\multirow{4}{*}{3 Views} & 1 & 2 & 3 & 4 & 5 & 6 & 7 & 8 & 9 & 10 & 11 & 12 \\
                   & 13.75 & 15.51 & 16.27 & 16.63 & 16.70 & 16.73 & 16.75 & 16.76 & 16.76 & 16.76 & 16.75 & 16.75 \\
                   \cline{2-13}
                   & 13 & 14 & 15 & 16 & 17 & 18 & 19 & 20 & 21 & 22 & 23 & 24 \\
                   & 16.75 & 16.75 & 16.75 & 16.74 & 16.74 & 16.74 & 16.73 & 16.73 & 16.73 & 16.73 & 16.72 & 16.72 \\
                   \hline
\multirow{4}{*}{6 Views} & 1 & 2 & 3 & 4 & 5 & 6 & 7 & 8 & 9 & 10 & 11 & 12 \\
                   & 14.63 & 16.28 & 17.63 & 18.21 & 18.55 & 18.76 & 18.89 & 18.97 & 19.02 & 19.05 & 19.06 & 19.08 \\
                   \cline{2-13}
                   & 13 & 14 & 15 & 16 & 17 & 18 & 19 & 20 & 21 & 22 & 23 & 24 \\
                   & 19.08 & 19.09 & 19.09 & 19.09 & 19.09 & 19.09 & 19.09 & 19.09 & 19.08 & 19.08 & 19.08 & 19.08 \\
                   \hline
\multirow{4}{*}{9 Views} & 1 & 2 & 3 & 4 & 5 & 6 & 7 & 8 & 9 & 10 & 11 & 12 \\
                   & 14.30 & 17.09 & 18.80 & 19.66 & 20.12 & 20.51 & 20.81 & 20.97 & 21.06 & 21.12 & 21.16 & 21.21 \\
                   \cline{2-13}
                   & 13 & 14 & 15 & 16 & 17 & 18 & 19 & 20 & 21 & 22 & 23 & 24 \\
                   & 21.25 & 21.26 & 21.28 & 21.29 & 21.31 & 21.32 & 21.33 & 21.34 & 21.34 & 21.34 & 21.34 & 21.35 \\
                   \hline
\end{tabular}
\captionsetup{width=.8\textwidth}
\caption{Average results in PSNR (dB) for each slimmed rank of SlimmeRF-24 in sparse-view scenarios. In each grid the number on the top is the number of components left and the number on the bottom is the average PSNR value.}
\label{LLFFavg}
\end{table*}

\FloatBarrier
\newpage

\FloatBarrier
\newpage

{\small
\bibliographystyle{ieee_fullname}

}